\documentclass{article}

\usepackage[T1]{fontenc}

\usepackage[bitstream-charter]{mathdesign}
\usepackage{amsmath}
\usepackage[scaled=0.92]{PTSans}

\usepackage[
  paper  = letterpaper,
  left   = 1.25in,
  right  = 1.25in,
  top    = 1.0in,
  bottom = 1.0in,
  ]{geometry}
\usepackage{setspace}

\usepackage[usenames,dvipsnames]{xcolor}
\definecolor{shadecolor}{gray}{0.9}

\usepackage[english]{babel}
\usepackage[parfill]{parskip}
\usepackage{afterpage}
\usepackage{framed}
\usepackage{nicefrac}

\DeclareRobustCommand{\parhead}[1]{\textbf{#1}~}

\usepackage{graphicx}
\usepackage[labelfont=bf]{caption}
\usepackage[format=hang]{subcaption}

\usepackage{booktabs}

\usepackage{natbib}

\usepackage[algoruled]{algorithm2e}

\usepackage[acronym,smallcaps,nowarn]{glossaries}

\usepackage[colorlinks,bookmarks=false]{hyperref}
\hypersetup{citecolor=Violet}
\hypersetup{linkcolor=black}
\hypersetup{urlcolor=MidnightBlue}

\newcommand{\gray}[1]{\textcolor{black!60}{#1}}

\definecolor{POSTcolor}{rgb}{0.48, 0.20, 0.58} 
\definecolor{Qcolor}{rgb}{0.00, 0.53, 0.22}  
\DeclareRobustCommand{\mb}[1]{\mathbf{\boldsymbol{#1}}}

\DeclareRobustCommand{\KL}[2]{\ensuremath{\text{KL}\left(#1\;\|\;#2\right)}}

\DeclareMathOperator*{\argmax}{arg\,max}
\DeclareMathOperator*{\argmin}{arg\,min}

\renewcommand{\mid}{~\vert~}

\newcommand{\g}{\mid}

\newcommand{\mbw}{\mb{w}}

\newcommand{\mbx}{\mb{x}}

\newcommand{\mby}{\mb{y}}

\newcommand{\mbz}{\mb{z}}

\newcommand{\mbI}{\mb{I}}
\newcommand{\mbone}{\mb{1}}

\newcommand{\mbL}{\mb{L}}

\newcommand{\mbtheta}{\mb{\theta}}

\newcommand{\mbomega}{\mb{\omega}}

\newcommand{\mbsigma}{\mb{\sigma}}
\newcommand{\mbSigma}{\mb{\Sigma}}
\newcommand{\mbphi}{\mb{\phi}}
\newcommand{\mbPhi}{\mb{\Phi}}

\newcommand{\mbalpha}{\mb{\alpha}}
\newcommand{\mbbeta}{\mb{\beta}}

\newcommand{\mbeta}{\mb{\eta}}
\newcommand{\mbmu}{\mb{\mu}}
\newcommand{\mbrho}{\mb{\rho}}

\newcommand{\mbzeta}{\mb{\zeta}}

\newcommand\dif{\mathop{}\!\mathrm{d}}
\newcommand{\diag}{\text{diag}}
\newcommand{\supp}{\text{supp}}

\newcommand{\E}{\mathbb{E}}
\newcommand{\bbH}{\mathbb{H}}

\newcommand{\bbR}{\mathbb{R}}

\newcommand{\cL}{\mathcal{L}}

\newcommand{\cN}{\mathcal{N}}

\newcommand{\Gam}{\text{Gam}}
\newcommand{\InvGam}{\text{InvGam}}

\newacronym{KL}{kl}{Kullback-Leibler}
\newacronym{ELBO}{elbo}{evidence lower bound}

\newacronym{MCMC}{mcmc}{Markov chain Monte Carlo}
\newacronym{MC}{mc}{Monte Carlo}

\newacronym{VI}{vi}{variational inference}
\newacronym{SVI}{svi}{stochastic variational inference}

\newacronym{ADVI}{advi}{automatic differentiation variational inference}
\newacronym{BBVI}{bbvi}{black box variational inference}

\newacronym{RMSPROP}{rmsprop}{rmsprop}

\newacronym{NUTS}{nuts}{no-U-turn sampler}
\newacronym{HMC}{hmc}{Hamiltonian Monte Carlo}

\newacronym{ARD}{ard}{automatic relevance determination}
\newacronym{GMM}{gmm}{Gaussian mixture model}
\newacronym{LDA}{lda}{latent Dirichlet allocation}

\newacronym{SGA}{sga}{stochastic gradient ascent}

\newacronym{PPCA}{ppca}{probabilistic principal component analysis}
\newacronym{SUP-PPCA}{sup-ppca}{supervised probabilistic principal component
analysis}

\makeglossaries

\begin{document}

\title{\textbf{Automatic Differentiation Variational Inference}}

\author{\textbf{Alp Kucukelbir}\\
Data Science Institute, Department of Computer Science \\ 
Columbia University\\ \\
\textbf{Dustin Tran}\\
Department of Computer Science \\ 
Columbia University\\ \\
\textbf{Rajesh Ranganath}\\
Department of Computer Science \\ 
Princeton University\\ \\
\textbf{Andrew Gelman}\\
Data Science Institute, Departments of Political Science and Statistics \\ 
Columbia University\\ \\
\textbf{David M.~Blei}\\
Data Science Institute, Departments of Computer Science and Statistics\\ 
Columbia University
}

\maketitle

\begin{abstract}  Probabilistic modeling is iterative.  A scientist posits a simple
  model, fits it to her data, refines it according to her analysis,
  and repeats.  However, fitting complex models to large data is a
  bottleneck in this process.  Deriving algorithms for new models can
  be both mathematically and computationally challenging, which makes
  it difficult to efficiently cycle through the steps.  To this end,
  we develop \gls{ADVI}.  Using our method, the scientist only
  provides a probabilistic model and a dataset, nothing else.
  \gls{ADVI} automatically derives an efficient variational inference
  algorithm, freeing the scientist to refine and explore many models.
  \gls{ADVI} supports a broad class of models---no conjugacy
  assumptions are required.  We study \gls{ADVI} across ten different
  models and apply it to a dataset with millions of observations.
  \gls{ADVI} is integrated into Stan, a probabilistic programming
  system; it is available for immediate use.
\end{abstract}

\glsreset{ADVI}

\section{Introduction}

We develop an automatic method that derives variational
inference algorithms for complex probabilistic models.  We implement
our method in Stan, a probabilistic programming system that lets a
user specify a model in an intuitive programming language and then
compiles that model into an inference executable.  Our method
enables fast inference with large datasets on an expansive class of
probabilistic models.

Figure\nobreakspace \ref {fig:porto_ppca} gives an example. Say we want to analyze how
people navigate a city by car. We have a dataset of all the taxi rides
taken over the course of a year: 1.7 million trajectories. To explore
patterns in this data, we propose a mixture model with an unknown
number of components. This is a non-conjugate model that we seek to
fit to a large dataset. Previously, we would have to manually derive
an inference algorithm that scales to large data. In our method, we
write a Stan program and compile it; we can then fit the model
in minutes and analyze the results with ease.

The context of this research is the field of probabilistic modeling,
which has emerged as a powerful language for customized data
analysis. Probabilistic modeling lets us express our assumptions about
data in a formal mathematical way, and then derive algorithms that use
those assumptions to compute about an observed dataset.  It has had an
impact on myriad applications in both statistics and machine learning,
including natural language processing, speech recognition, computer
vision, population genetics, and computational neuroscience.

Probabilistic modeling leads to a natural research cycle.  A scientist
first uses her domain knowledge to posit a simple model that includes
latent variables; then, she uses an inference algorithm to infer
those variables from her data; next, she analyzes her results and
identifies where the model works and where it falls short; last, she
refines the model and repeats the process.  When we cycle through
these steps, we find expressive, interpretable, and useful models
\citep{gelman2013bayesian,blei2014build}. One of the broad goals of 
machine learning is to make this process easy.

Looping around this cycle, however, is not easy.  The data we study
are often large and complex; accordingly, we want to propose rich
probabilistic models and scale them up.  But using such models
requires complex algorithms that are difficult to derive, implement,
and scale.  The bottleneck is this computation 
that precludes the scientist from taking full advantage of
the probabilistic modeling cycle.

This problem motivates the important ideas of probabilistic
programming and automated inference.  Probabilistic programming allows
a user to write a probability model as a computer program and then
compile that program into an efficient inference
executable. Automated inference is the backbone of such a system---it
inputs a probability model, expressed as a program, and outputs an
efficient algorithm for computing with it.  Previous approaches to
automatic inference have mainly relied on \gls{MCMC}
algorithms.  The results have been successful, but automated 
\gls{MCMC} is too slow for many real-world applications.

We approach the problem through variational inference, a faster
alternative to \gls{MCMC} that has been used in many large-scale
problems
\citep{blei2016variational}.
Though it is a promising method, developing a variational inference
algorithm still requires tedious model-specific derivations and
implementation; it has not seen widespread use in probabilistic
programming.  Here we automate the process of deriving scalable
variational inference algorithms.  We build on recent ideas in
so-called ``black-box'' variational inference to leverage strengths of
probabilistic programming systems, namely the ability to transform the
space of latent variables and to automate derivatives of the joint
distribution. The result, called \gls{ADVI}, provides an automated
solution to variational inference: the inputs are a probabilistic
model and a dataset; the outputs are posterior inferences about the
model's latent variables.\footnote{This paper extends the method presented in \citep
{kucukelbir2015automatic}.}
We implemented and deployed \gls{ADVI} as
part of Stan, a probabilistic programming system
\citep{stan-manual:2015}.

\gls{ADVI} in Stan resolves the computational bottleneck of the
probabilistic modeling cycle. A scientist can easily propose a
probabilistic model, analyze a large dataset, and revise the model,
without worrying about computation.
\gls{ADVI} enables this cycle by providing automated and scalable 
variational inference for an expansive class of models.
Sections\nobreakspace \ref {sec:empirical} and\nobreakspace  \ref {sec:practice} present ten probabilistic modeling examples,
including a progressive analysis of 1.7 million taxi trajectories.

\parhead{Technical summary.}  Formally, a probabilistic model defines
a joint distribution of observations $\mbx$ and latent variables
$\mbtheta$, $p(\mbx, \mbtheta)$.  The inference problem is to compute
the posterior, the conditional distribution of the latent variables
given the observations $p(\mbtheta \mid \mbx)$.  The
posterior can reveal patterns in the data and
form predictions through the posterior predictive distribution.
The problem is that, for many models, the posterior is not tractable
to compute.

Variational inference turns the task of computing a posterior into
an optimization problem. We posit a parameterized family of
distributions $q(\mbtheta) \in \mathcal{Q}$ and then find the member
of that family that minimizes the \gls{KL} divergence to the exact
posterior.  Traditionally, using a variational inference algorithm
requires the painstaking work of developing and implementing a custom
optimization routine: specifying a variational family appropriate to
the model, computing the corresponding objective function, taking
derivatives, and running a gradient-based or coordinate-ascent
optimization.

\gls{ADVI} solves this problem automatically.  The user specifies the
model, expressed as a program, and \gls{ADVI} automatically generates
a corresponding variational algorithm.  The idea is to first
automatically transform the inference problem into a common space and
then to solve the variational optimization. Solving the problem in
this common space solves variational inference for all models in a
large class.  In more detail, \gls{ADVI} follows these steps.

\begin{enumerate}

\item \gls{ADVI} transforms the model into one with
  unconstrained real-valued latent variables. 
  Specifically, it  transforms $p(\mbx, \mbtheta)$ to
  $p(\mbx, \mbzeta)$, where the mapping from $\mbtheta$ to $\mbzeta$
  is built into the joint distribution. 
  This removes all original constraints on the latent variables 
  $\mbtheta$.  \gls{ADVI} then defines the
  corresponding variational problem on the transformed variables, that
  is, to minimize $\KL{q(\mbzeta)}{p(\mbzeta \mid \mbx)}$.  With this
  transformation, all latent variables are defined on the
  same space. \gls{ADVI} can now use a single variational family for 
  all models.

\item \gls{ADVI} recasts the
  gradient of the variational objective function as an expectation
  over $q$. This involves the gradient of the
  log joint with respect to the latent variables
  $\nabla_{\mbtheta} \log p(\mbx, \mbtheta)$.  Expressing the
  gradient as an expectation opens the door to Monte Carlo methods
  for approximating it \citep{robert1999monte}.

\item \gls{ADVI} further reparameterizes the gradient in terms
  of a standard Gaussian. To do this, it uses another transformation,
  this time within the variational family.  This second transformation
  enables \gls{ADVI} to efficiently compute Monte Carlo approximations---it
  needs only to sample from a standard Gaussian 
  \citep{kingma2013auto,rezende2014stochastic}.

\item \gls{ADVI} uses noisy gradients to
  optimize the variational distribution \citep{robbins1951stochastic}.
  An adaptively tuned step-size sequence provides
  good convergence in practice \citep{bottou2012stochastic}.

\end{enumerate}

We developed \gls{ADVI} in the Stan system, which gives us two
important types of automatic computation around probabilistic models.
First, Stan provides a library of transformations---ways to convert a
variety of constrained latent variables (e.g., positive reals) to be
unconstrained, without changing the underlying joint
distribution. Stan's library of transformations helps us with
step 1 above. Second, Stan implements automatic
differentiation to calculate $\nabla_\mbtheta \log p(\mbx, \mbtheta)$
\citep{carpenter2015stan,baydin2015automatic}.
These derivatives are crucial in step 2, when computing the
gradient of the \gls{ADVI} objective.

\parhead{Organization of paper.} Section\nobreakspace \ref {sec:advi} develops the recipe
that makes \gls{ADVI}. We expose the details of each of the steps
above and present a concrete algorithm. Section\nobreakspace \ref {sec:empirical} studies
the properties of \gls{ADVI}.  We explore its accuracy, its stochastic
nature, and its sensitivity to transformations. Section\nobreakspace \ref {sec:practice}
applies \gls{ADVI} to an array of probability models. We compare its
speed to \gls{MCMC} sampling techniques and present a case study using
a dataset with millions of observations.  Section\nobreakspace \ref {sec:discussion}
concludes the paper with a discussion.

\section{Automatic Differentiation Variational Inference}
\label{sec:advi}
\glsreset{ADVI}

\Gls{ADVI} offers a recipe for automating the
computations involved in variational inference. The strategy is as
follows: transform the latent variables of the model into a common
space, choose a variational approximation in the common space, and use
generic computational techniques to solve the variational problem.

\subsection{Differentiable Probability Models}
\label{sub:diff_prob_models}

We begin by defining the class of probability models that \gls{ADVI}
supports.  Consider a dataset $\mbx = x_{1:N}$ with $N$
observations. Each $x_n$ is a realization of a discrete or continuous
(multivariate) random variable. The likelihood $p (\mbx\mid\mbtheta)$ relates
the observations to a set of latent random variables $\mbtheta$. A
Bayesian model posits a prior density $p(\mbtheta)$ on the latent
variables. Combining the likelihood with the prior gives the joint
density $p(\mbx,\mbtheta) = p(\mbx\mid\mbtheta)\,p(\mbtheta)$.  The
goal of inference is to compute the posterior density
$p (\mbtheta\mid\mbx)$, which describes how the latent variables vary,
conditioned on data.

Many posterior densities are not tractable to compute; their normalizing
constants lack analytic (closed-form) solutions.  Thus we often seek
to approximate the posterior.  \gls{ADVI} approximates the
posterior of differentiable probability models.  Members of this
class of models have continuous latent variables $\mbtheta$ and a
gradient of the log-joint with respect to them,
$\nabla_\mbtheta \log p(\mbx,\mbtheta)$. The gradient is valid within
the support of the prior
\begin{align*}
  \supp(p(\mbtheta))
  &=
  \big\{\,
  \mbtheta \mid \mbtheta \in \bbR^K
  \text{ and }
  p(\mbtheta) > 0
  \,\big\} \subseteq \bbR^{K},
\end{align*}
where $K$ is the dimension of the latent variable space. This support
set is important: it will
play a role later in the paper. We make no assumptions about
conjugacy, either full \citep{diaconis1979conjugate} or conditional
\citep{hoffman2013stochastic}.

Consider a model that contains a Poisson likelihood with unknown rate,
$p(x\mid\theta)$.  The observed variable $x$ is discrete; the latent
rate $\theta$ is continuous and positive.  Place a Weibull prior on
$\theta$, defined over the positive real numbers. The resulting joint
density describes a nonconjugate differentiable probability model; the
posterior distribution of $\theta$ is not in the same class as the
prior.  (The conjugate prior would be a Gamma.)  However, it is in the
class of differentiable models.  Its partial derivative
$\nicefrac{\partial}{\partial\theta} \, \log p (x,\theta)$ is valid within the
support of the Weibull distribution,
$\supp(p(\theta)) = \bbR_{>0} \subset \bbR$. While this model would be
a challenge for classical variational inference, it is not for
\gls{ADVI}.

Many machine learning models are differentiable. For example: linear
and logistic regression, matrix factorization with continuous or
discrete observations, linear dynamical systems, and Gaussian
processes.  (See Table\nobreakspace \ref {tab:diff_nondiff_models}.)
At first blush, the restriction to continuous random
variables may seem to leave out other common machine learning models,
such as mixture models, hidden Markov models, and topic models.
However, marginalizing out the discrete variables in the likelihoods
of these models renders them differentiable.\footnote{
  Marginalization is not tractable for all models, such as the Ising
  model, sigmoid belief network, and (untruncated) Bayesian
  nonparametric models, such as Dirichlet process mixtures
  \citep{blei2006variational}.  These are not differentiable
  probability models.}  

\begin{table}[h]
\centering
\begin{tabular}{ll}
{Generalized linear models} &
\gray{\emph{(e.g.,~linear / logistic / probit)}} \\
{Mixture models} &
\gray{\emph{(e.g.,~mixture of Gaussians)}} \\
{Deep exponential families} &
\gray{\emph{(e.g.,~deep latent Gaussian models)} } \\
{Topic models} &
\gray{\emph{(e.g.,~latent Dirichlet allocation)} }  \\
{Linear dynamical systems} &
\gray{\emph{(e.g.,~state space models, hidden Markov models)} }  \\
{Gaussian process models} &
\gray{\emph{(e.g.,~regression / classification)} }
\end{tabular}
\caption{Popular differentiable probability models in machine learning.}
\label{tab:diff_nondiff_models}
\end{table}

\glsreset{VI}
\subsection{Variational Inference}
\label{sub:variational_inference}

Variational inference turns approximate posterior inference into an
optimization
problem \citep{wainwright2008graphical,blei2016variational}.
Consider a family of approximating densities of the latent variables
$q (\mbtheta\,;\,\mbphi)$, parameterized by a vector 
$\mbphi \in \mbPhi$.  \Gls{VI} finds the parameters that minimize the
\gls{KL} divergence to the posterior,
\begin{align}
  \mbphi^*
  &=
  \argmin_{\mbphi\in\mbPhi} \KL{q (\mbtheta\,;\,\mbphi)}{p(\mbtheta\mid\mbx)}.
  \label{eq:min_KL}
\end{align}
The optimized $q(\mbtheta \, ; \, \mbphi^*)$ then serves as an approximation to
the posterior.

The \gls{KL} divergence lacks an analytic form because it involves the
posterior. Instead we maximize the \gls {ELBO}
\begin{align}
  \cL (\mbphi)
  &=
  \E_{q (\mbtheta)} \big[ \log p (\mbx,\mbtheta) \big]
  -
  \E_{q (\mbtheta)} \big[ \log q (\mbtheta\,;\,\mbphi) \big].
  \label{eq:elbo}
\end{align}
The first term is an expectation of the joint density under the
approximation, and the second is the entropy of the variational
density.  The \gls{ELBO} is equal to the negative \gls{KL} divergence
up to the constant $\log p(\mbx)$.  Maximizing the \gls{ELBO}
minimizes the \gls{KL} divergence
\citep{jordan1999introduction,bishop2006pattern}.

Optimizing the \gls{KL} divergence implies a constraint that the
support of the approximation $q(\mbtheta \, ; \, \mbphi)$ lie within 
the support of the posterior $p(\mbtheta \g \mbx)$.\footnote{If $\supp(q) \not\subseteq \supp (p)$ then outside the support of $p$
we have $\KL{q}{p} = \E_q [\log q] - \E_q[\log p] = -\infty$.} 
With this constraint made explicit, the optimization problem from
Equation\nobreakspace \textup {(\ref {eq:min_KL})} becomes
\begin{align}
  \mbphi^*
  &=
  \argmax_{\mbphi\in\mbPhi} \cL(\mbphi)
  \quad\text{such that}\quad
  \supp(q(\mbtheta\,;\,\mbphi)) \subseteq \supp(p(\mbtheta \mid \mbx)).
  \label{eq:variational_problem_latent_variable_space}
\end{align}
We explicitly include this constraint because we have not specified
the form of the variational approximation; we must ensure that
$q(\mbtheta\,;\,\mbphi)$ stays within the support of the posterior.

The support of the posterior, however, may also be unknown. So, we
further assume that the support of the posterior equals that of the
prior, $\supp(p(\mbtheta \mid \mbx)) = \supp(p(\mbtheta))$.  This is a
benign assumption, which holds for most models considered in
machine learning.  In detail, it holds when the likelihood does not
constrain the prior; i.e., the likelihood must be positive over the
sample space for any $\mbtheta$ drawn from the prior.

\smallskip
\parhead{Our recipe for automating \gls{VI}.} The traditional way of
solving Equation\nobreakspace \textup {(\ref {eq:variational_problem_latent_variable_space})} is
difficult.  We begin by choosing a variational family
$q(\mbtheta\,;\,\mbphi)$ that, by definition, satisfies the support
matching constraint.  We compute the expectations in the \gls {ELBO},
either analytically or through approximation. We then decide on a
strategy to maximize the \gls{ELBO}. For instance, we might use
coordinate ascent by iteratively updating the components of $\mbphi$.
Or, we might follow gradients of the \gls{ELBO} with respect to
$\mbphi$ while staying within $\mbPhi$. Finally, we implement, test,
and debug software that performs the above.  Each step requires expert
thought and analysis in the service of a single algorithm for a single
model.

In contrast, our approach allows the scientist to define any
differentiable probability model, and we automate the process of
developing a corresponding \gls{VI} algorithm.  Our recipe for
automating \gls{VI} has three ingredients.  First, we automatically
transform the support of the latent variables $\mbtheta$ to the real
coordinate space (Section\nobreakspace \ref {sub:auto_transform_constrained}); 
this lets us choose from a variety of variational
distributions $q$ without worrying about the support matching
constraint (Section\nobreakspace \ref {sub:variational_approximation_in_real}).   
Second, we compute the \gls{ELBO} for any model using
\gls{MC} integration, which only requires being able to sample from
the variational distribution (Section\nobreakspace \ref {sub:variational_probem_in_real_coord}). 
Third, we employ stochastic gradient
ascent to maximize the \gls{ELBO} and use automatic differentiation to
compute gradients without any user input (Section\nobreakspace \ref {sub:stochastic_optimization}). 
With these tools, we can
develop a generic method that automatically solves the variational
optimization problem for a large class of models.

\subsection{Automatic Transformation of Constrained Variables}
\label{sub:auto_transform_constrained}

We begin by transforming the support of the latent variables
$\mbtheta$ such that they live in the real coordinate space $\bbR^K$.
Once we have transformed the density, we can choose the variational
approximation independent of the model.

Define a one-to-one differentiable function
\begin{align}
  T &: \text{supp}(p(\mbtheta)) \rightarrow \bbR^K,
  \label{eq:transformation}
\end{align}
and identify the transformed variables as $\mbzeta = T(\mbtheta)$.
The transformed joint density $p(\mbx,\mbzeta)$ is a function of $\mbzeta$; it
has the representation
\begin{align*}
  p(\mbx,\mbzeta)
  &=
  p
  \left(\mbx,T^{-1}(\mbzeta)\right)
  \big|
  \det J_{T^{-1}}(\mbzeta)
  \big|,
\end{align*}
where $p(\mbx,\mbtheta = T^{-1}(\mbzeta))$ is the joint density in the original
latent variable space,
and $J_{T^{-1}}(\mbzeta)$ is the Jacobian of the inverse of
$T$. Transformations of continuous probability densities require a
Jacobian; it accounts for how the transformation warps unit volumes
and ensures that the transformed density integrates to one~\citep
{olive2014statistical}. (See Appendix\nobreakspace \ref {app:jacobian}.)

\begin{figure}[tb]
\hspace*{-0.15in}
\includegraphics{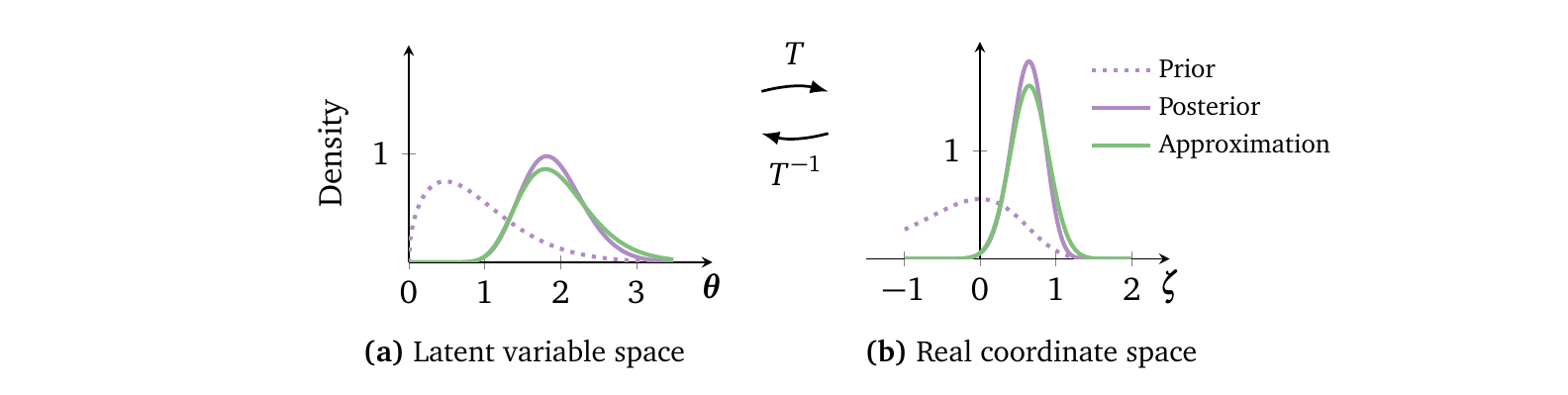}
\caption{Transforming the latent variable to real coordinate space.
The \textcolor {POSTcolor} {purple} line is the posterior.
The \textcolor {Qcolor}{green} line is the approximation.
\textbf{(a)}
The latent variable space is $\bbR_{>0}$.
\textbf{(a$\to$b)}
$T$ transforms the latent variable space to $\bbR$.
\textbf{(b)}
The variational approximation is a Gaussian in real coordinate space.}
\label{fig:transformations_AB}
\end{figure}

Consider again our running Weibull-Poisson example from
Section\nobreakspace \ref {sub:diff_prob_models}.  
The latent variable $\theta$
lives in $\bbR_{>0}$. The logarithm $\zeta = T(\theta) = \log(\theta)$
transforms $\bbR_{>0}$ to the real line $\bbR$. Its Jacobian
adjustment is the derivative of the inverse of the logarithm
$|\det J_{T^{-1}(\zeta)}| = \exp(\zeta)$.  The transformed density is
\begin{align*}
p(x,\zeta) =
\text{Poisson}(x\mid\exp(\zeta))
\times
\text{Weibull}(\exp(\zeta)\:;\:1.5,1)
\times
\exp(\zeta).
\end{align*}
Figure\nobreakspace \ref {fig:transformations_AB} depicts this transformation.

\begin{figure}[tbp]
\hspace*{0.25in}
\includegraphics{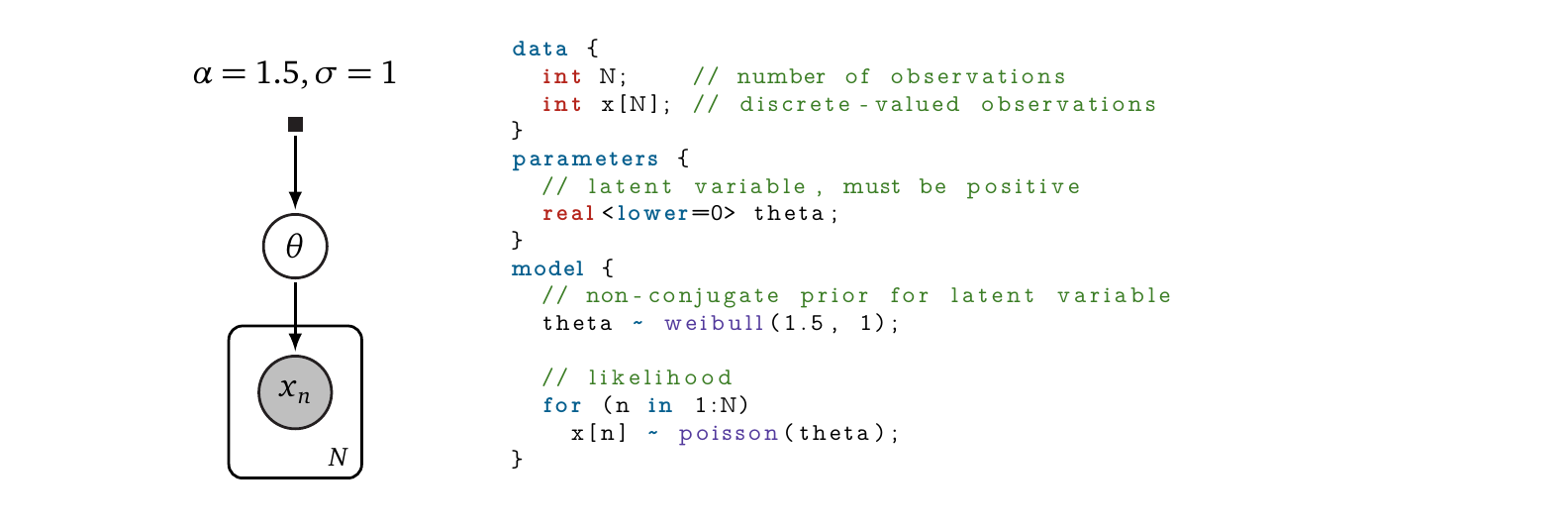}
\caption{Specifying a simple nonconjugate probability model in Stan.}
\label{fig:example_poisson}
\end{figure}

As we describe in the introduction, we implement our algorithm in
Stan~\citep{stan-manual:2015}.  Stan maintains a library of
transformations and their corresponding Jacobians.\footnote{Stan
  provides various transformations for upper and lower bounds, simplex
  and ordered vectors, and structured matrices such as covariance
  matrices and Cholesky factors.}  With Stan, we can automatically
transforms the joint density of any differentiable probability model
to one with real-valued latent variables. (See Figure\nobreakspace \ref {fig:example_poisson}.)

\subsection{Variational Approximations in Real Coordinate Space}
\label{sub:variational_approximation_in_real}

After the transformation, the latent variables $\mbzeta$ have support
in the real coordinate space $\bbR^K$. We have a choice of variational
approximations in this space. Here, we consider Gaussian distributions; 
these implicitly induce non-Gaussian variational
distributions in the original latent variable space.

\parhead{Mean-field Gaussian.}
One option is to posit a factorized (mean-field) Gaussian variational
approximation
\begin{align*}
  q(\mbzeta \,;\, \mbphi)
  &=
  \cN\left(\mbzeta \,;\, \mbmu, \diag(\mbsigma^2)\right)
  =
  \prod_{k=1}^K
  \cN
  \left(\zeta_k \,;\, \mu_k, \sigma^2_k\right),
\end{align*}
where the vector
$\mbphi = (\mu_{1},\cdots,\mu_{K}, \sigma^2_ {1},\cdots,\sigma^2_{K})$
concatenates the mean and variance of each Gaussian factor. Since the variance
parameters must always be positive, the variational parameters live in the set
$\mbPhi = \{\bbR^K, \bbR_{>0}^K\}$. 
Re-parameterizing the mean-field Gaussian removes this constraint.
Consider the logarithm of the standard deviations, $\mbomega = \log(\mbsigma)$,
applied
element-wise. The support of $\mbomega$ is now the real coordinate
space and $\mbsigma$ is always positive.
The mean-field Gaussian becomes
$
q(\mbzeta \,;\, \mbphi)
=
\cN\left(\mbzeta \,;\, \mbmu, \diag(\exp(\mbomega)^2)\right),
$
where the vector
$\mbphi = (\mu_{1},\cdots,\mu_{K}, \omega_ {1},\cdots,\omega_{K})$
concatenates the mean and logarithm of the standard deviation of each factor.
Now, the variational parameters are unconstrained in $\bbR^{2K}$.

\parhead{Full-rank Gaussian.}
Another option is to posit a full-rank Gaussian variational approximation
\begin{align*}
  q(\mbzeta \,;\, \mbphi)
  &=
  \cN\left(\mbzeta \,;\, \mbmu, \mbSigma\right),
\end{align*}
where the vector
$\mbphi = (\mbmu, \mbSigma)$ concatenates the mean vector $\mbmu$
and covariance matrix $\mbSigma$.
To ensure that $\mbSigma$ always remains positive semidefinite, we
re-parameterize the covariance matrix using a Cholesky factorization,
$\mbSigma = \mbL\mbL^\top$.
We use the non-unique definition of the Cholesky factorization
where the diagonal elements of $\mbL$ need not be positively constrained \citep
{pinheiro1996unconstrained}. Therefore $\mbL$ lives in the unconstrained space
of lower-triangular matrices with $K(K+1)/2$ real-valued entries.
The full-rank Gaussian becomes
$
q(\mbzeta \,;\, \mbphi)
=
\cN\left(\mbzeta \,;\, \mbmu, \mbL\mbL^\top\right),
$
where the variational parameters $\mbphi = (\mbmu,\mbL)$ are unconstrained in
$\bbR^{K+K(K+1)/2}$.

The full-rank Gaussian generalizes the mean-field Gaussian approximation. The
off-diagonal terms in the covariance matrix $\mbSigma$ capture
posterior correlations across latent random variables.\footnote{This is a form of structured mean-field variational inference \citep
{wainwright2008graphical,barber2012bayesian}.}
This leads to a more accurate posterior
approximation than the mean-field Gaussian; however, it comes at a
computational cost. Various low-rank approximations to the covariance
matrix reduce this cost, yet limit its ability to model complex
posterior correlations \citep{seeger2010gaussian,challis2013gaussian}.

\parhead{The choice of a Gaussian.}
Choosing a Gaussian distribution may call to mind the Laplace
approximation technique, where a second-order Taylor expansion around
the maximum-a-posteriori estimate gives a Gaussian approximation to
the posterior.  However, using a Gaussian variational approximation is
not equivalent to the Laplace approximation
\citep{opper2009variational}. Our approach is distinct in another way:
the posterior approximation in the original latent variable space
is non-Gaussian.

\parhead{The implicit variational density.}
The transformation $T$ from Equation\nobreakspace \textup {(\ref {eq:transformation})} maps the support of the
latent variables to the real coordinate space. Thus, its inverse $T^{-1}$ maps
back to the support of the latent variables. This implicitly defines the
variational approximation in the original latent variable space as
$
q \left(T(\mbtheta)\,;\, \mbphi\right)
\big| \det J_{T}(\mbtheta) \big|.
$
The transformation ensures that the support of this approximation is always
bounded by that of the posterior in the original latent variable space.

\parhead{Sensitivity to $T$.}
There are many ways to transform the support a
variable to the real coordinate space. The form of the
transformation directly affects the shape of the variational
approximation in the original latent variable space.  
We study sensitivity to the choice of transformation
in Section\nobreakspace \ref {sub:sensitivity_to_transformations}.

\subsection{The Variational Problem in Real Coordinate Space}
\label{sub:variational_probem_in_real_coord}

Here is the story so far. We began with a differentiable probability model 
$p(\mbx,\mbtheta)$. We transformed the latent variables into $\mbzeta$, which
live in the real coordinate space. We defined variational approximations in the
transformed space. Now, we consider the variational optimization problem.

Write the variational objective function, the \gls{ELBO}, in real
coordinate space as
\begin{align}
  \cL(\mbphi)
  &=
  \E_{q(\mbzeta\,;\,\mbphi)}
  \bigg[
  \log p \left(\mbx, T^{-1}(\mbzeta)\right)
  +
  \log \big| \det J_{T^{-1}}(\mbzeta) \big|
  \bigg]
  +
  \bbH
  \big[
  q(\mbzeta\,;\,\mbphi)
  \big].
  \label{eq:elbo_real_coordinate_space}
\end{align}
The inverse of the transformation $T^{-1}$ appears in the joint model, along
with the determinant of the Jacobian adjustment. The \gls{ELBO} is a function of
the variational parameters $\mbphi$ and the entropy $\bbH$, both of which depend
on the variational approximation.  (Derivation in Appendix\nobreakspace \ref {app:elbo_unconstrained}.)

Now, we can freely optimize the \gls{ELBO} in the
real coordinate space without worrying about the support
matching constraint. The optimization problem from
Equation\nobreakspace \textup {(\ref {eq:variational_problem_latent_variable_space})}
becomes
\begin{align}
  \mbphi^*
  &=
  \argmax_{\mbphi} \cL(\mbphi)
  \label{eq:unconst_variational_problem_real_coordinate_space}
\end{align}
where the parameter vector $\mbphi$ lives in some appropriately dimensioned real
coordinate space. This is an unconstrained optimization problem that we
can solve using gradient ascent. Traditionally, this would require manual
computation of gradients. Instead, we develop a stochastic gradient ascent
algorithm that uses automatic differentiation to compute gradients and \gls{MC}
integration to approximate expectations.

We cannot directly use automatic differentiation on the \gls{ELBO}. This is
because the \gls{ELBO} involves an unknown expectation. However, we can
automatically differentiate the functions inside the expectation. (The model
$p$ and transformation $T$ are both easy to represent as computer functions
\citep{baydin2015automatic}.)
To apply automatic differentiation, we want to push the gradient operation
inside the expectation. To this end, we employ one final transformation:
elliptical standardization\footnote{Also known as
a ``coordinate transformation'' \citep{rezende2014stochastic},
an ``invertible transformation'' \citep{titsias2014doubly},
and
the ``re-parameterization trick'' \citep{kingma2013auto}.}
\citep{hardle2012applied}.

\parhead{Elliptical standardization.}
Consider a transformation $S_\mbphi$ that absorbs the variational parameters
$\mbphi$; this converts the Gaussian variational approximation into a standard
Gaussian. In the mean-field case, the standardization is
$
\mbeta = S_{\mbphi}(\mbzeta) =
\diag\left(\exp\left(\mbomega\right)\right)^{-1} (\mbzeta - \mbmu)
$. In the full-rank Gaussian, the standardization is
$
\mbeta = S_{\mbphi}(\mbzeta) =
\mbL^{-1} (\mbzeta - \mbmu)
$.

\begin{figure}[htb]
\hspace*{-0.15in}
\includegraphics{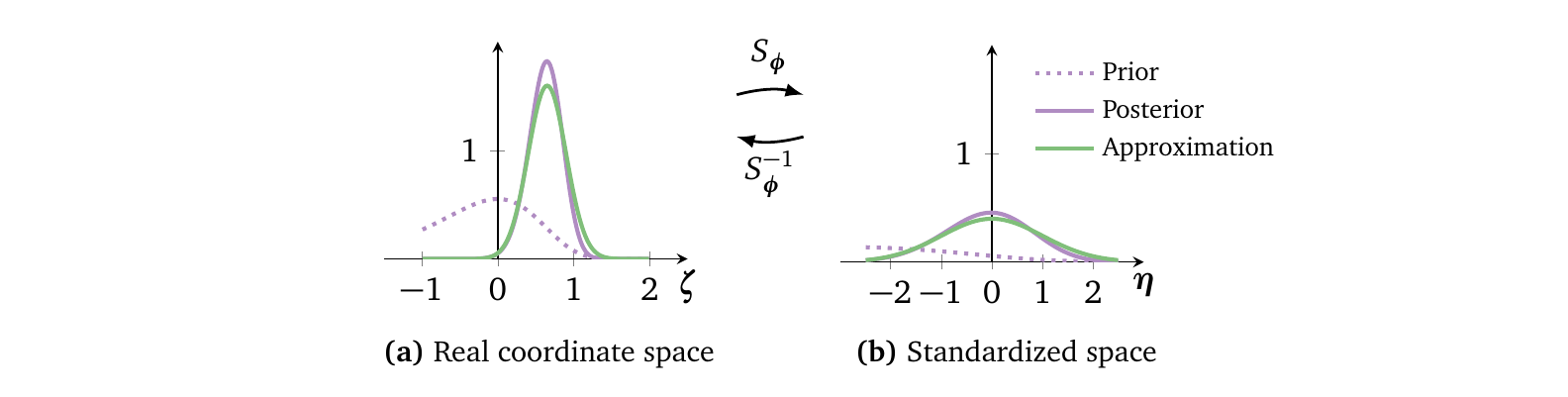}
\caption{Elliptical standardization.
The \textcolor {POSTcolor} {purple} line is the posterior.
The \textcolor {Qcolor}{green} line is the approximation.
\textbf{(a)}
The variational approximation is a Gaussian in real coordinate space.
\textbf{(a$\to$b)}
$S_{\mbphi}$ absorbs the parameters of the Gaussian.
\textbf{(b)}
We maximize the \gls{ELBO} in the standardized space, with a fixed
approximation. The \textcolor {Qcolor} {green} line is a standard
Gaussian. }
\label{fig:transformations_BC}
\end{figure}

In both cases, the standardization encapsulates the variational parameters; in
return it gives a fixed variational density
\begin{align*} q (\mbeta)
  &=
  \cN
  \left(\mbeta \,;\, \mb{0}, \mbI\right)
  =
  \prod_{k=1}^K
  \cN
  \left(\eta_k \,;\, 0, 1\right),
\end{align*}
as shown in Figure\nobreakspace \ref {fig:transformations_BC}.

The standardization transforms the variational problem from
Equation\nobreakspace \textup {(\ref {eq:elbo_real_coordinate_space})} into
\begin{align*}
\mbphi^*
  &=
  \argmax_{\mbphi}
  \E_{\cN(\mbeta\,;\, \mb{0}, \mbI)}
  \bigg[
  \log p
  \left(\mbx,T^{-1}(S_{\mbphi}^{-1}(\mbeta))\right)
  +
  \log \big| \det J_{T^{-1}}\left(S_{\mbphi}^{-1}(\mbeta)\right) \big|
  \bigg]
  +
  \bbH
  \big[
  q(\mbzeta\,;\,\mbphi)
  \big].
\end{align*}
The expectation is now in terms of a standard Gaussian density.
The Jacobian of elliptical standardization evaluates to one, because the
Gaussian distribution is a member of the location-scale family: standardizing a
Gaussian gives another Gaussian distribution. (See Appendix\nobreakspace \ref {app:jacobian}.)

We do not need to transform the entropy term as it does not depend on the model
or the transformation; we have a simple analytic form for the entropy of a
Gaussian and its gradient. We implement these once and reuse for all models.

\subsection{Stochastic Optimization}
\label{sub:stochastic_optimization}

We now reach the final step: stochastic optimization of the variational
objective function.

\parhead{Computing gradients.}
Since the expectation is no longer dependent on $\mbphi$, we can directly
calculate its gradient. Push the gradient inside the expectation and apply the
chain rule to get
\begin{align}
  \nabla_\mbmu \cL
  &=
  \E_{\cN(\mbeta)}
  \left[
  \nabla_\mbtheta \log p(\mbx,\mbtheta)
  \nabla_{\mbzeta} T^{-1}(\mbzeta)
  +
  \nabla_{\mbzeta}
  \log \big| \det J_{T^{-1}}(\mbzeta) \big|
  \right].
  \label{eq:elbo_grad_mu}
\end{align}

We obtain gradients with respect to $\mbomega$ (mean-field) and $\mbL$
(full-rank) in a similar fashion
\begin{align}
  \nabla_{\mbomega} \cL
  &=
  \E_{\cN(\mbeta)}
  \left[
  \left(
  \nabla_{\mbtheta} \log p(\mbx,\mbtheta)
  \nabla_{\mbzeta} T^{-1}(\mbzeta)
  +
  \nabla_{\mbzeta}
  \log \big| \det J_{T^{-1}}(\mbzeta) \big|
  \right)
  \mbeta^\top
  \diag(\exp(\mbomega))
  \right]
  + \mbone
  \label{eq:elbo_grad_omega}\\
  \nabla_{\mbL}
  \cL
  &=
  \E_{\cN(\mbeta)}
  \left[
  \left(
  \nabla_\mbtheta \log p(\mbx,\mbtheta)
  \nabla_{\mbzeta} T^{-1}(\mbzeta)
  +
  \nabla_{\mbzeta}
  \log \big| \det J_{T^{-1}}(\mbzeta) \big|
  \right)
  \mbeta^\top
  \right] + (\mbL^{-1})^\top.
  \label{eq:elbo_grad_L}
\end{align}
(Derivations in Appendix\nobreakspace \ref {app:gradient_elbo}.)

We can now compute the gradients inside the expectation with automatic
differentiation. The only thing left is the expectation. \gls
{MC} integration provides a simple approximation: draw samples
from the standard Gaussian and evaluate the empirical mean of the
gradients within the expectation (Appendix\nobreakspace \ref {app:mc_integration}). In practice a
single sample suffices. (We study this in detail in 
Section\nobreakspace \ref {sub:variance_of_gradient_estimators} and in the experiments in 
Section\nobreakspace \ref {sec:practice}.)

This gives noisy unbiased gradients of the \gls{ELBO} for any differentiable
probability model. We can now use these gradients in a stochastic optimization
routine to automate variational inference.

\parhead{Stochastic gradient ascent.}
Equipped with noisy unbiased gradients of the \gls{ELBO}, \gls{ADVI} implements
stochastic gradient ascent (Algorithm\nobreakspace \ref {alg:ADVI}). This algorithm
is guaranteed to converge to a local maximum of the \gls{ELBO} under certain
conditions on the step-size sequence.\footnote{This is also called a \emph{learning rate} or \emph{schedule} in the
machine learning community.}
Stochastic gradient ascent falls under the class of stochastic approximations,
where
\citet{robbins1951stochastic} established a pair of conditions that ensure
convergence: most prominently, the step-size sequence must decay sufficiently
quickly. Many sequences satisfy these criteria, but their specific forms impact
the success of stochastic gradient ascent in practice. We describe an adaptive
step-size sequence for \gls{ADVI} below.

\parhead{Adaptive step-size sequence.}
Adaptive step-size sequences retain (possibly infinite) memory about past
gradients and adapt to the high-dimensional curvature of the \gls{ELBO}
optimization space
\citep{amari1998natural,duchi2011adaptive,ranganath2013adaptive,kingma2014adam}. 
These sequences enjoy
theoretical bounds on their convergence rates. However, in practice, they can be
slow to converge. The empirically justified \textsc{rmsprop} sequence
\citep{rmsprop}, which only retains finite memory of past gradients, converges
quickly in practice but lacks any convergence guarantees. We propose a new
step-size sequence which effectively combines both approaches.

Consider the step-size $\mbrho^{(i)}$ and a gradient vector $\mb{g}^{(i)}$
at iteration $i$. We define the $k$th element of $\mbrho^{(i)}$ as
\begin{align}
  \rho_k^{(i)}
  &=
  \eta
  \times
  i^{-\nicefrac{1}{2}+\epsilon}
  \times
  \left(\tau + \sqrt{s^{(i)}_k}\right)^{-1},
  \label{eq:stepsize}
\end{align}
where we apply the following recursive update
\begin{align}
  s^{(i)}_k
  &=
  \alpha {g^2_k}^{(i)} + (1-\alpha) s^{(i-1)}_k,
  \label{eq:rmsprop}
\end{align}
with an initialization of $s^{(1)}_k = {g^2_k}^{(1)}$.

The first factor $\eta \in \bbR_{>0}$ controls the scale of the step-size sequence.
It mainly affects the beginning of the optimization. We adaptively tune $\eta$
by searching over $\eta \in \{0.01, 0.1, 1, 10, 100\}$ using a
subset of the data and selecting the value that leads to the fastest convergence
\citep{bottou2012stochastic}.

The middle term $i^{-1/2+\epsilon}$ decays as a function of the iteration $i$.
We set $\epsilon=10^{-16}$, a small value that guarantees that the step-size
sequence satisfies the \citet{robbins1951stochastic} conditions.

The last term adapts to the curvature of the \gls{ELBO} optimization space.
Memory about past gradients are processed in Equation\nobreakspace \textup {(\ref {eq:rmsprop})}. The weighting
factor $\alpha\in(0,1)$  defines a compromise of old and new gradient
information, which we set to $0.1$. The quantity $s_k$
converges to a non-zero constant. Without the previous decaying term, this would
lead to possibly large oscillations around a local optimum of the \gls{ELBO}.
The additional perturbation $\tau > 0$ prevents division by zero and down-weights
early iterations. In practice the step-size is not very sensitive to this value
\citep{hoffman2013stochastic}, so we set $\tau=1$.

\parhead{Complexity and data subsampling.}
\gls{ADVI} has complexity $\mathcal{O}(NMK)$ per iteration, where $N$ is the
number of data points, $M$ is the number of \gls{MC} samples (typically between
1 and 10), and $K$ is the number of latent variables. Classical
\gls{VI} which hand-derives a coordinate ascent algorithm has complexity 
$\mathcal{O}(NK)$ per pass over the dataset. The added complexity of automatic
differentiation over analytic gradients is roughly constant 
\citep{carpenter2015stan,baydin2015automatic}.

\begin{algorithm}[t]
  \caption{Automatic Differentiation Variational Inference}
  \SetAlgoLined
  \DontPrintSemicolon
  \BlankLine
  \KwIn{Dataset $\mbx=x_{1:N}$, model $p(\mbx,\mbtheta)$.}
  Set iteration counter $i = 1$.\;
  Initialize $\mbmu^{(1)} = \mb{0}$.\;
  Initialize $\mbomega^{(1)} = \mb{0}$ (mean-field) or
             $\mbL^{(1)} = \mbI$ (full-rank).\;
  Determine $\eta$ via a search over finite values.\;
  \BlankLine
  \While{change in \gls{ELBO} is above some threshold}{
    \BlankLine
    Draw $M$ samples $\mbeta_m \sim \cN(\mb{0},\mbI)$ from
    the standard multivariate Gaussian.\;
    \BlankLine
    Approximate $\nabla_\mbmu \cL$ using \gls{MC} integration
    (Equation\nobreakspace \textup {(\ref {eq:elbo_grad_mu})}).\;
    \BlankLine
    Approximate $\nabla_{\mbomega} \cL$ or $\nabla_{\mbL} \cL$
    using \gls{MC} integration
    (Equations\nobreakspace \textup {(\ref {eq:elbo_grad_omega})} and\nobreakspace  \textup {(\ref {eq:elbo_grad_L})}).\;
    \BlankLine
    Calculate step-size $\mbrho^{(i)}$ (Equation\nobreakspace \textup {(\ref {eq:stepsize})}).\;
    \BlankLine
    Update
    $\mbmu^{(i+1)}            \longleftarrow
    \mbmu^{(i)} + \diag(\mbrho^{(i)})\nabla_\mbmu \cL$.\;
    \BlankLine
    Update
    $\mbomega^{(i+1)}  \longleftarrow
    \mbomega^{(i)}  + \diag(\mbrho^{(i)})\nabla_{\mbomega} \cL$
    or
    $\mbL^{(i+1)}  \longleftarrow
    \mbL^{(i)}  + \diag(\mbrho^{(i)})\nabla_{\mbL} \cL$.\;
    \BlankLine
    Increment iteration counter.\;
  }
  Return
  $\mbmu^* \longleftarrow \mbmu^{(i)}$.\;
  \BlankLine
  Return
  $\mbomega^* \longleftarrow \mbomega^{(i)}$
  or
  $\mbL^* \longleftarrow \mbL^{(i)}$.\;
  \BlankLine
  \label{alg:ADVI}
\end{algorithm}

We scale  \gls{ADVI} to large datasets using stochastic optimization
with data subsampling
\citep{hoffman2013stochastic,titsias2014doubly}. The adjustment to
Algorithm\nobreakspace \ref {alg:ADVI} is simple:
sample a minibatch of size $B \ll N$ from the dataset and scale the likelihood
of the model by $N/B$ \citep{hoffman2013stochastic}.
The stochastic extension of \gls{ADVI} has a per-iteration complexity $\mathcal
{O}(BMK)$.

In Sections\nobreakspace \ref {sub:gmm} and\nobreakspace  \ref {sub:taxicab}, we apply this stochastic extension to analyze
datasets with millions of observations.

\subsection{Related Work}
\gls{ADVI} automates variational inference
within the Stan probabilistic programming system. 
This draws on two major themes.

The first theme is probabilistic programming. One class of systems
focuses on probabilistic models where the user specifies a joint
probability distribution. Some examples are BUGS 
\citep{spiegelhalter1995bugs}, JAGS \citep{plummer2003jags}, and 
Stan \citep{stan-manual:2015}. Another class of systems allows the user
to directly specify more general probabilistic programs. Some examples are
Church \citep{goodman2008church}, Figaro \citep{pfeffer2009figaro},
Venture \citep{mansinghka2014venture}, and
Anglican \citep{wood2014new}. Both classes primarily rely on various
forms of \gls{MCMC} techniques for inference; they typically cannot
scale to very large data.

The second is a body of work that generalizes variational 
inference. \citet{ranganath2014black} and
\citet{salimans2014using} propose a black-box technique that only
requires computing gradients of the variational approximating family.
\citet{kingma2013auto} and \citet{rezende2014stochastic} describe a
reparameterization of the variational problem that simplifies
optimization.
\citet{titsias2014doubly} leverage the gradient of the model for a 
class of real-valued models. \citet{rezende2015variational} and
\citet{tran2015variational} improve the accuracy of
black-box variational approximations. Here we build on and extend these ideas to
automate variational inference; we highlight technical connections as we study
the properties of \gls{ADVI} in Section\nobreakspace \ref {sec:empirical}.

Some notable work crosses both themes.
\citet{bishop2002vibes} present an automated 
variational algorithm for graphical models with conjugate exponential
relationships between all parent-child pairs. \citet
{winn2005variational,InferNET14} extend this to graphical models with
non-conjugate relationships by either using custom approximations or an
expensive sampling approach.
\gls{ADVI} automatically supports a more comprehensive class of nonconjugate
models; see Section\nobreakspace \ref {sub:diff_prob_models}.
\citet{wingate2013automated} study a more general setting, where the 
variational approximation itself is a probabilistic program.

\section{Properties of Automatic Differentiation Variational Inference}
\label{sec:empirical}
\glsreset{ADVI}

\Gls{ADVI} extends classical variational inference techniques in a few
directions. 
In this section, we use simulated data to study three
aspects of \gls{ADVI}: the accuracy of mean-field and full-rank
approximations, the variance of the \gls{ADVI} gradient estimator, and
the sensitivity to the transformation $T$.

\subsection{Accuracy}
\label{sub:accuracy}

We begin by considering three models that expose how the mean-field
approximation affects the accuracy of \gls {ADVI}.

\parhead{Two-dimensional Gaussian.} We first study a simple model that
does not require approximate inference.  Consider a multivariate
Gaussian likelihood $\cN(\mby \mid\mbmu,\mbSigma)$ with fixed, yet highly
correlated,
covariance $\mbSigma$; our goal is to estimate the mean $\mbmu$.  If we
place a multivariate Gaussian prior on $\mbmu$ then the posterior is
also a Gaussian that we can compute analytically~\citep{bernardo2009bayesian}.

\begin{figure}[tbp]
\includegraphics{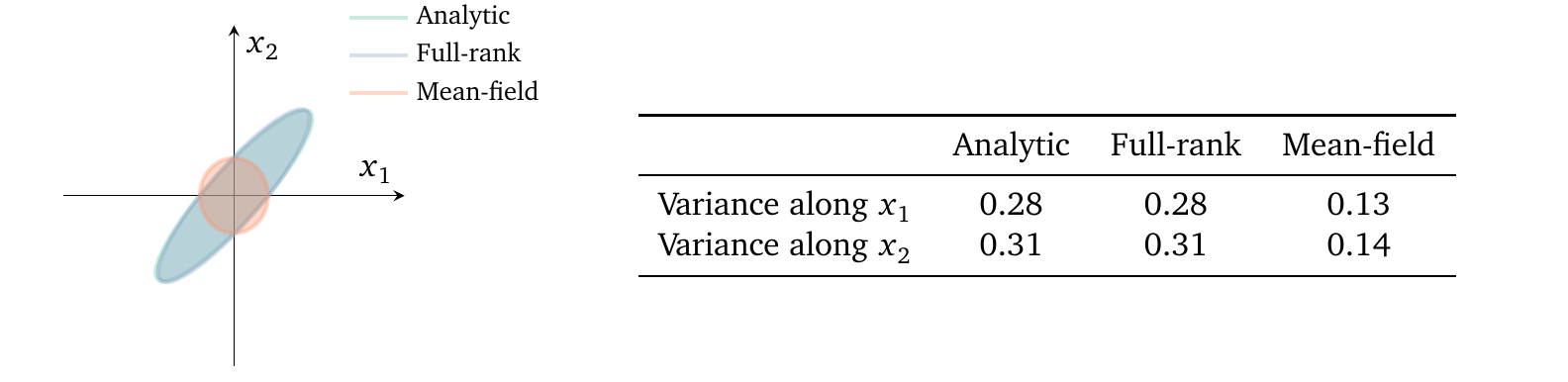}
  \caption{Comparison of mean-field and full-rank \gls{ADVI} on a
  two-dimensional Gaussian model. The figure shows the accuracy of the full-rank
  approximation. Ellipses correspond to two-sigma level sets of the Gaussian.
  The table quantifies the underestimation of marginal variances by the
  mean-field approximation. }
  \label{fig:underestimation}
\end{figure}

We draw $1000$ datapoints from the model and run both variants of \gls
{ADVI}, mean-field and full-rank, until
convergence. Figure\nobreakspace \ref {fig:underestimation} compares the \gls{ADVI}
methods to the exact posterior.  Both procedures correctly identify
the mean of the analytic posterior. However, the shape of the
mean-field approximation is incorrect.  This is because the mean-field
approximation ignores off-diagonal terms of the Gaussian
covariance. \gls{ADVI} minimizes the \gls{KL} divergence from the
approximation to the exact posterior; this leads to a systemic
underestimation of marginal variances \citep{bishop2006pattern}.

\parhead{Logistic regression.}  We now study a model for which we need
approximate inference.  Consider logistic regression, a generalized
linear model with a binary response $y$, covariates $\mbx$, and
likelihood
$\text{Bern} (y \mid \text{logit}^{-1} (\mbx^\top \mbbeta))$; our goal
is to estimate the coefficients $\mbbeta$. We place an independent
Gaussian prior on each regression coefficient.

\begin{figure}[htb]
\hspace*{-0.1in}
\includegraphics{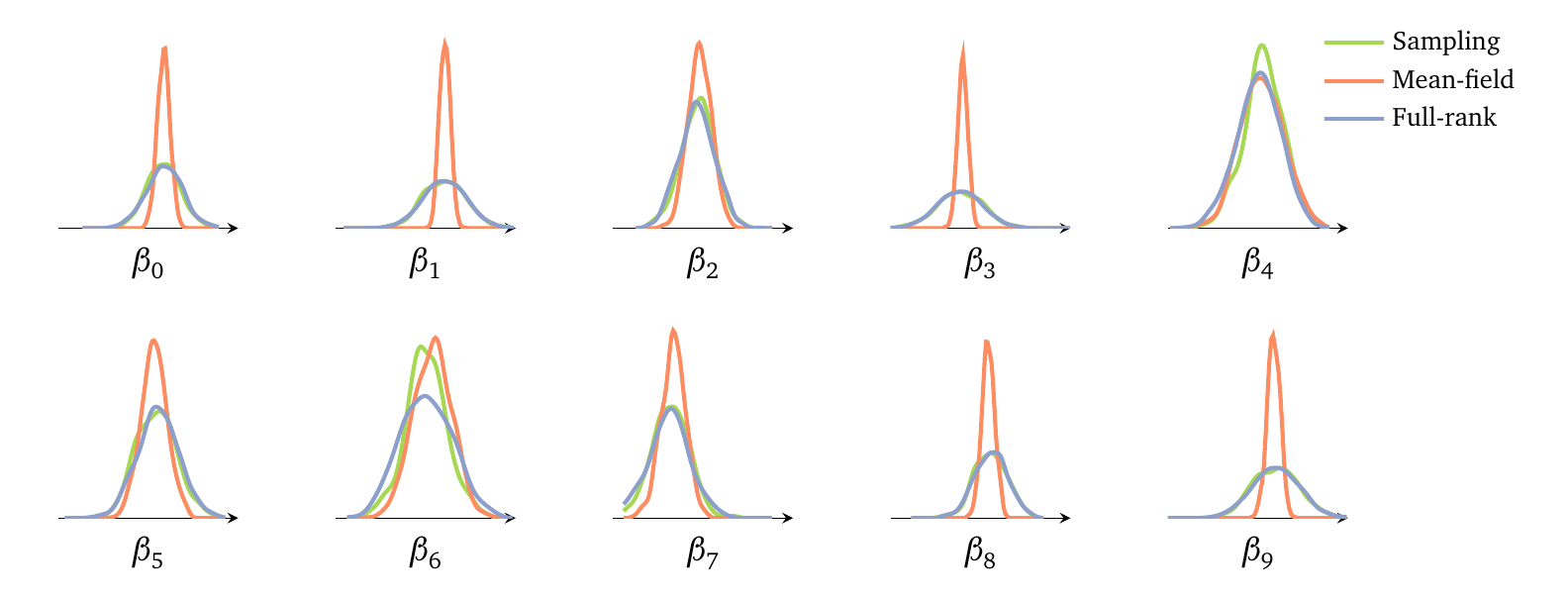}
\caption{Comparison of marginal posterior densities for a logistic regression
model. Each plot shows kernel density estimates for the posterior of each
coefficient using $1000$ samples. Mean-field \gls{ADVI} underestimates variances
for most of the coefficients.}
\label{fig:logistic_accuracy}
\end{figure}

We simulated $9$ random covariates from the prior distribution (plus a constant
intercept) and drew $1000$ datapoints from the likelihood. We
estimated the posterior of the coefficients with \gls{ADVI} and Stan's default
\gls{MCMC} technique, the \gls{NUTS} \citep{hoffman2014nuts}.  
Figure\nobreakspace \ref {fig:logistic_accuracy} shows the marginal posterior
densities obtained from each approximation.  \gls{MCMC} and
\gls{ADVI} perform similarly in their estimates of the posterior
mean. The mean-field approximation, as expected, underestimates
marginal posterior variances on most of the coefficients. The
full-rank approximation, once again, better matches the posterior.

\parhead{Stochastic volatility time-series model.}  Finally, we study
a model where the data are not exchangeable. Consider an
autoregressive process to model how the latent volatility (i.e.,
variance) of an economic asset changes over
time~\citep{kim1998stochastic}; our goal is to estimate the sequence
of volatilities. We expect these posterior estimates to be
correlated, especially when the volatilities trend away from their mean value.

In detail, the price data exhibit latent volatility as part of the
variance of a zero-mean Gaussian
\begin{align*}
  y_t \sim \cN\left(0,\exp(h_t/2)\right)
\end{align*}
where the log volatility follows an auto-regressive process
\begin{align*}
  h_t \sim \cN\left(\mu + \phi(h_{t-1}-\mu),\sigma\right)
  \quad\text{with initialization}\quad
  h_1 \sim \cN\left(\mu, \frac{\sigma}{\sqrt{1-\phi^2}}\right).
\end{align*}
We place the following priors on the latent variables
\begin{align*}
  \mu \sim \text{Cauchy}(0,10)
  ,\quad
  \phi \sim \text{Unif}(-1,1)
  ,\quad\text{and}\quad
  \sigma \sim \text{LogNormal}(0,10).
\end{align*}

\begin{figure}[htb]
\includegraphics{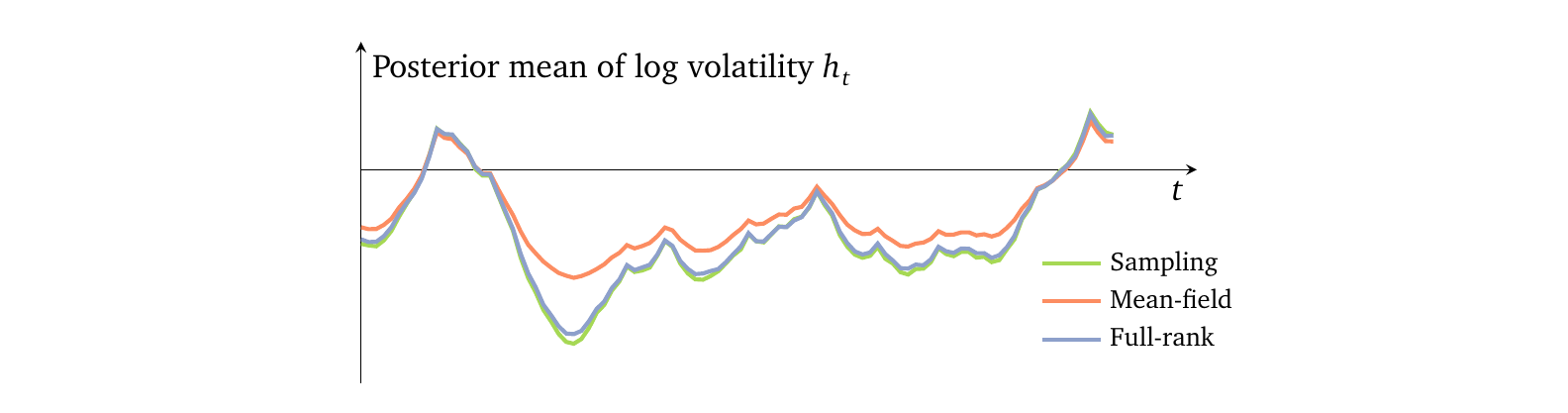}
\caption{Comparison of posterior mean estimates of volatility $h_t$. Mean-field
\gls{ADVI} underestimates $h_t$, especially when it moves far away from its
mean $\mu$. Full-rank \gls{ADVI} matches the accuracy of sampling.}
\label{fig:stocvol_plot}
\end{figure}

We set $\mu=-1.025$, $\phi=0.9$ and $\sigma=0.6$, and simulate a dataset of
$500$ time-steps from the generative model above. Figure\nobreakspace \ref {fig:stocvol_plot}
plots the posterior mean of the log volatility $h_t$ as a function of
time. Mean-field \gls{ADVI} struggles to describe the mean of the
posterior, particularly when the log volatility drifts far away from
$\mu$. In contrast, full-rank \gls {ADVI} matches the estimates
obtained from sampling.

\begin{figure}[htb]
\centering
\begin{subfigure}[t]{1.5in}
  \includegraphics[height=2.0in]{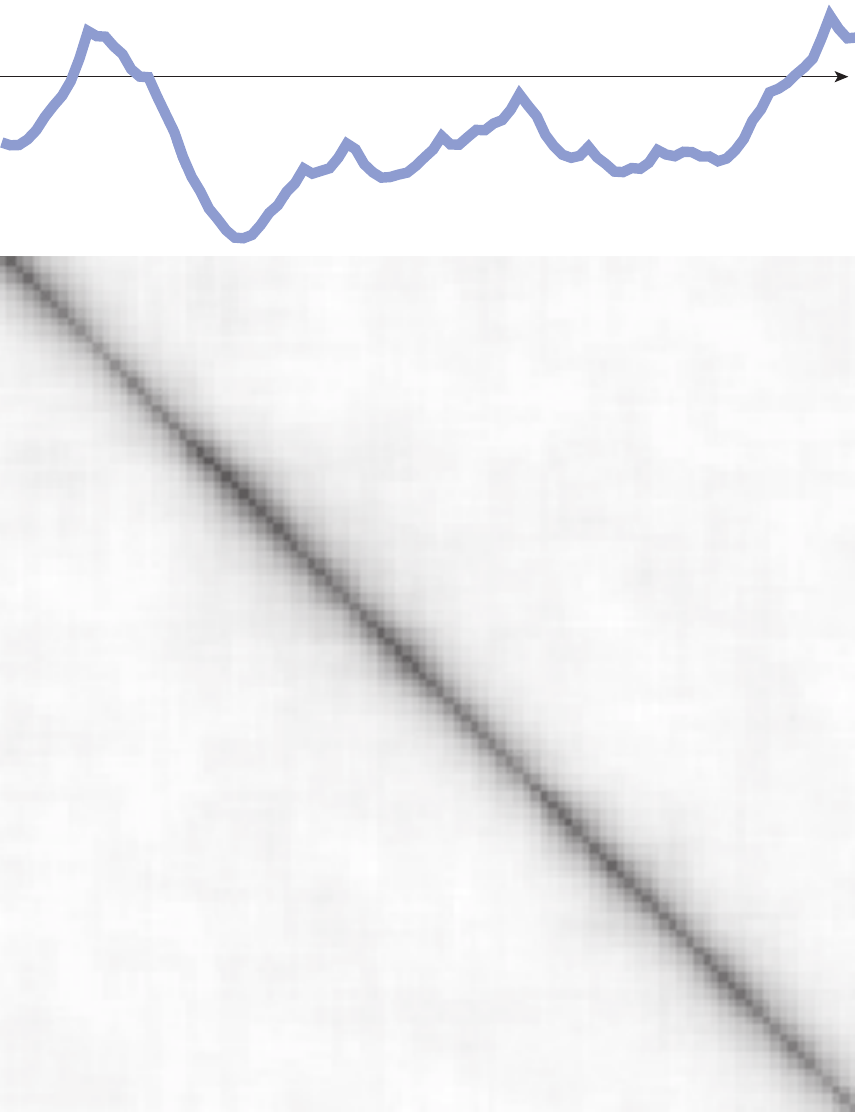}
  \caption{Mean-field}
  \label{sub:stocvol_mf}
\end{subfigure}
\hspace*{0.1in}
\begin{subfigure}[t]{1.5in}
  \includegraphics[height=2.0in]{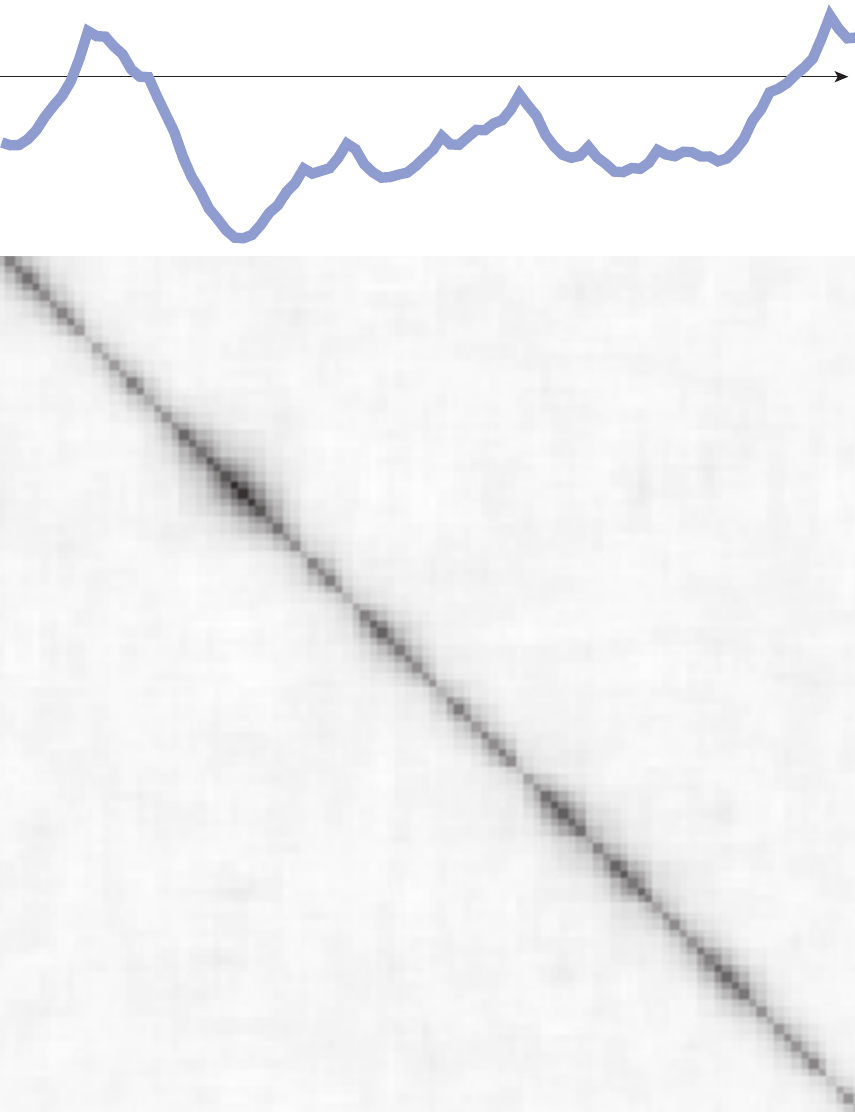}
  \caption{Full-rank}
  \label{sub:stocvol_fr}
\end{subfigure}
\hspace*{0.1in}
\begin{subfigure}[t]{1.5in}
  \includegraphics[height=2.0in]{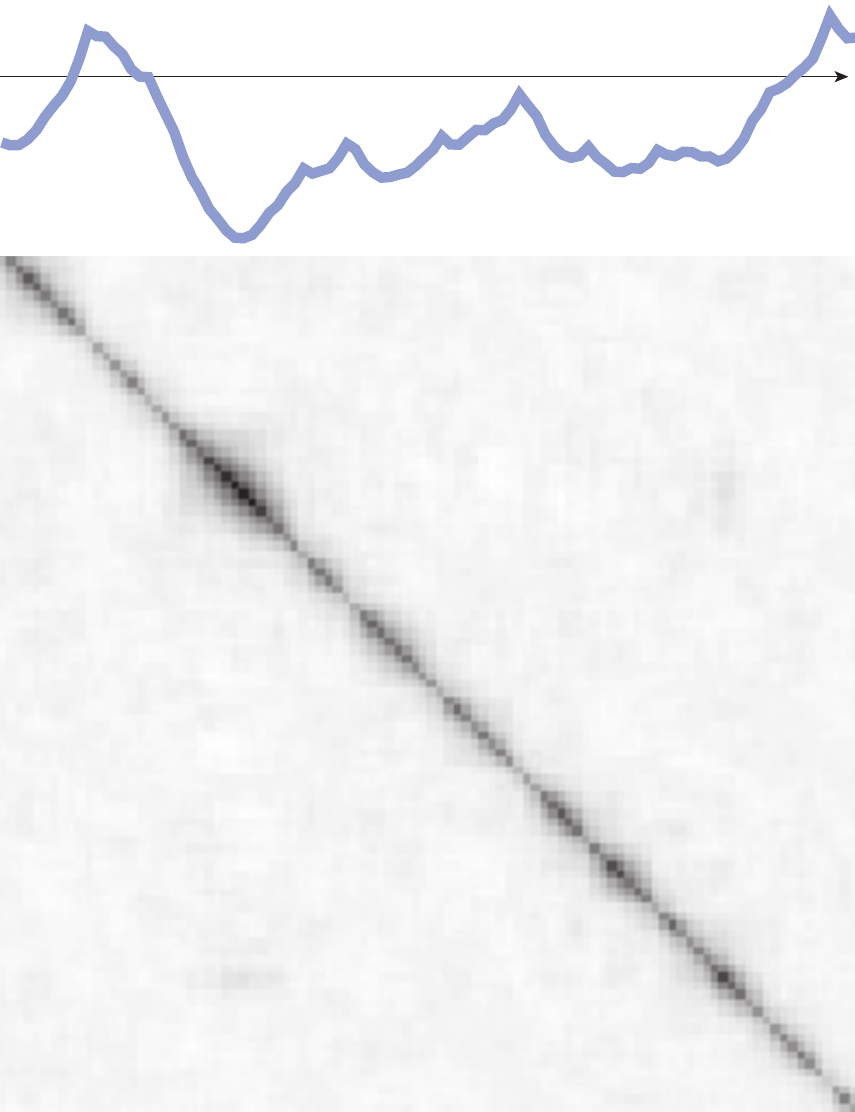}
  \caption{Sampling}
  \label{sub:stocvol_nuts}
\end{subfigure}
\hspace*{0.1in}
\begin{subfigure}[b]{0.25in}
  \includegraphics[height=1.55in]{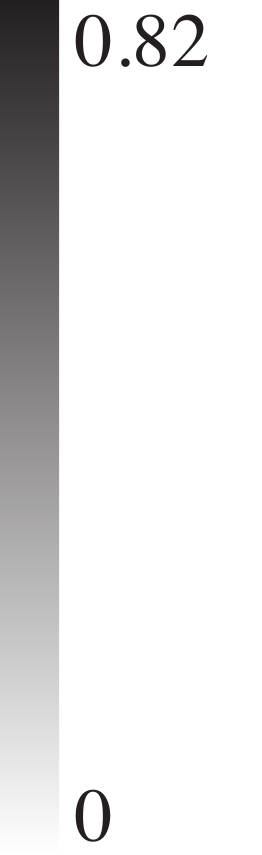}
\end{subfigure}
\caption{Comparison of empirical posterior covariance matrices. The mean-field
\gls{ADVI} covariance matrix fails to capture the local correlation structure
seen in the full-rank \gls{ADVI} and sampling results. All covariance matrices
exhibit a blurry spread due to finite sample size.}
\label{fig:stocvol_covariance}
\end{figure}

We further investigate this by studying posterior correlations of the
log volatility sequence. We draw $S=1000$ samples of $500$-dimensional log
volatility sequences $\{\mb{h}^{(s)}\}_1^{S}$.
Figure\nobreakspace \ref {fig:stocvol_covariance} shows the empirical posterior covariance matrix,
$\nicefrac{1}{S-1} \sum_s (\mb{h}^{(s)} - \overline{\mb{h}})
(\mb{h}^{(s)} - \overline{\mb{h}})^\top$
for each method. The mean-field covariance (fig.\nobreakspace \ref {sub:stocvol_mf}) 
fails to capture the locally correlated structure of the
full-rank and sampling covariance matrices
(figs.\nobreakspace \ref {sub:stocvol_fr} and\nobreakspace  \ref {sub:stocvol_nuts}). All covariance matrices
exhibit a blurry spread due to finite sample size.

The regions where the local correlation is strongest correspond to the regions
where mean-field underestimates the log volatility. To help identify these
regions, we overlay the sampling mean log volatility estimate from 
Figure\nobreakspace \ref {fig:stocvol_plot} above each matrix. Both full-rank \gls{ADVI} and
sampling results exhibit correlation where the log volatility trends away from
its mean value.

\parhead{Recommendations.}  How to choose between full-rank and
mean-field \gls{ADVI}?  Scientists interested in posterior variances
and covariances should use the full-rank approximation. Full-rank
\gls{ADVI} captures posterior correlations, in turn producing more
accurate marginal variance estimates.  For large data, however,
full-rank \gls{ADVI} can be prohibitively slow.

Scientists interested in prediction should initially rely on the
mean-field approximation.  Mean-field \gls{ADVI} offers a fast
algorithm for approximating the posterior mean. In practice, accurate
posterior mean estimates dominate predictive accuracy; underestimating
marginal variances matters less.

\subsection{Variance of the Stochastic Gradients}
\label{sub:variance_of_gradient_estimators}

\gls{ADVI} uses Monte Carlo integration to approximate gradients
of the \gls{ELBO}, and then uses these gradients in a stochastic
optimization algorithm (Section\nobreakspace \ref {sec:advi}).  The speed of \gls{ADVI}
hinges on the variance of the gradient estimates.  When a stochastic
optimization algorithm suffers from high-variance gradients, it must
repeatedly recover from poor parameter estimates.

\gls{ADVI} is not the only way to compute Monte Carlo approximations
of the gradient of the \gls{ELBO}.  \Gls{BBVI} takes a different
approach \citep{ranganath2014black}.  The \gls{BBVI} gradient
estimator uses the gradient of the variational approximation and
avoids using the gradient of the model.  For example, the following
\gls{BBVI} estimator
\begin{align*}
  \nabla_\mbmu^\textsc{bbvi} \cL
  &=
  \E_{q(\mbzeta\,;\,\mbphi)}
  \left[
  \nabla_\mbmu \log q(\mbzeta\,;\,\mbphi)
  \left\{
  \log p \left(\mbx, T^{-1}(\mbzeta)\right)
  +
  \log \big| \det J_{T^{-1}}(\mbzeta) \big|
  -
  \log q(\mbzeta\,;\,\mbphi)
  \right\}
  \right]
\end{align*}
and the \gls{ADVI} gradient estimator in Equation\nobreakspace \textup {(\ref {eq:elbo_grad_mu})} both lead
to unbiased estimates of the exact gradient. While \gls{BBVI} is more
general---it does not require the gradient of the model and thus
applies to more settings---its gradients can suffer from high variance.

\begin{figure}[htb]
\hspace*{-0.25in}
\includegraphics{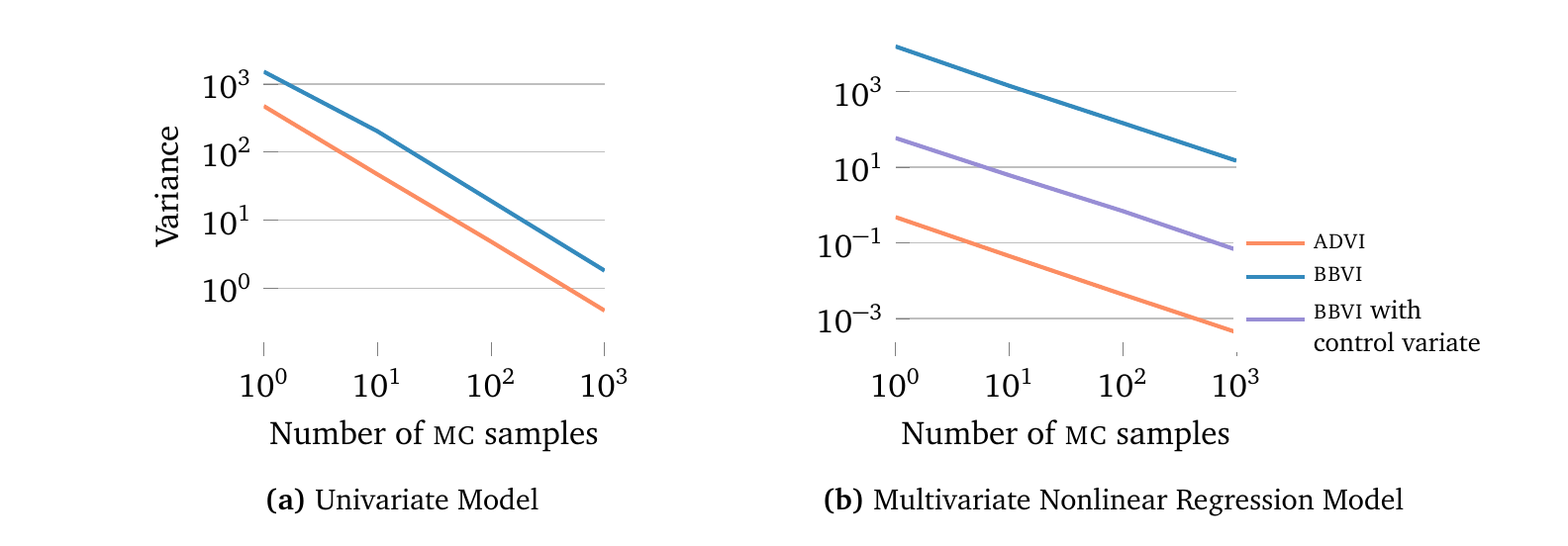}
\caption{Comparison of gradient estimator variances. The \gls{ADVI}
gradient estimator exhibits lower variance than the \gls{BBVI} estimator.
Moreover, it does not require control variate variance reduction, which is not
available in univariate situations.}
\label{fig:grad_variance} 
\end{figure}

Figure\nobreakspace \ref {fig:grad_variance} empirically compares the variance of both
estimators for two models. Figure\nobreakspace \ref {fig:grad_variance}a shows the
variance of both gradient estimators for a simple univariate model, where the
posterior is a $\text{Gamma}(10,10)$. We estimate the variance using ten
thousand re-calculations of the gradient $\nabla_\mbphi\cL$, across an
increasing number of \gls {MC} samples $M$. The \gls{ADVI} gradient has lower
variance; in practice, a single sample suffices. (See the experiments in
Section\nobreakspace \ref {sec:practice}.)

Figure\nobreakspace \ref {fig:grad_variance}b shows the same calculation for a
100-dimensional nonlinear regression model with likelihood
$\cN(\mby \mid \text{tanh}(\mbx^\top \mbbeta), \mbI)$ and a Gaussian prior on
the regression coefficients $\mbbeta$. Because this is a multivariate example,
we also show the \gls{BBVI} gradient with a variance
reduction scheme using control variates described in 
\citet{ranganath2014black}. In both cases, the \gls{ADVI} gradients are
statistically more efficient.

\subsection{Sensitivity to Transformations}
\label{sub:sensitivity_to_transformations}

\gls{ADVI} uses a transformation $T$ from the unconstrained space to
the constrained space.  We now study how the choice of this
transformation affects the non-Gaussian posterior approximation in the
original latent variable space.

\begin{figure}[htbp]
\hspace*{-0.25in}
\includegraphics{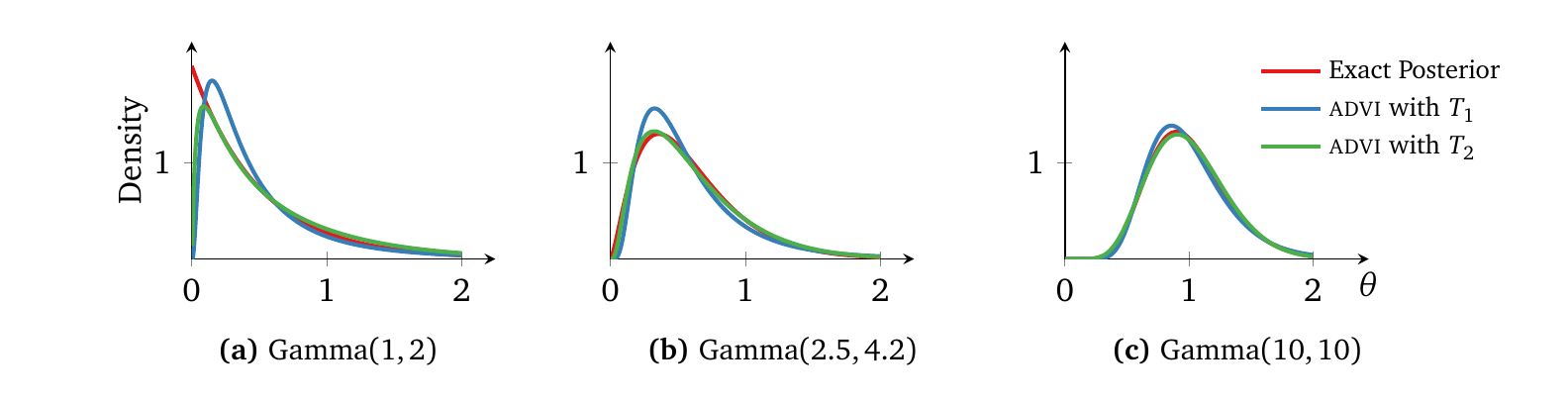}
\caption{\gls{ADVI} approximations to Gamma
densities under two different transformations.}
\label{fig:transform_gamma}
\end{figure}

Consider a posterior density in the Gamma family, with support over
$\bbR_{>0}$. Figure\nobreakspace \ref {fig:transform_gamma} shows three configurations of
the Gamma, ranging from $\text{Gamma}(1,2)$, which places most of its
mass close to $\theta=0$, to $\text{Gamma}(10,10)$, which is centered
at $\theta = 1$.  Consider two transformations $T_1$ and $T_2$
\begin{align*}
  T_1 : \theta\mapsto \log(\theta)
  \quad\text{and}\quad
  T_2 : \theta\mapsto \log(\exp(\theta) - 1),
\end{align*}
both of which map $\bbR_{>0}$ to $\bbR$. \gls{ADVI} can use either
transformation to approximate the Gamma posterior. Which one is
better?

Figure\nobreakspace \ref {fig:transform_gamma}
show the \gls{ADVI} approximation under both transformations. 
Table\nobreakspace \ref {tab:transform_gamma} reports the corresponding \gls{KL}
divergences. Both graphical and numerical results prefer $T_2$ over
$T_1$. A quick analysis corroborates this. $T_1$ is the logarithm,
which flattens out for large values.  However, $T_2$ is almost linear
for large values of $\theta$. Since both the Gamma (the posterior) and
the Gaussian (the \gls{ADVI} approximation) densities are
light-tailed, $T_2$ is the preferable transformation.

Is there an optimal transformation? Without
loss of generality, we consider fixing a standard Gaussian distribution in
the real coordinate space.\footnote{For two transformations $T_1$ and $T_2$ from latent variable
space to real coordinate space, there always exists a transformation $T_3$
within the real coordinate space such that $T_1(\theta) = T_3(T_2(\theta))$.}
The optimal transformation is then
\begin{align*}
  T^*
  &=
  \Phi^{-1} \circ P(\mbtheta \mid \mbx)
\end{align*}
where $P$ is the cumulative density function of the posterior and $\Phi^{-1}$
is the inverse cumulative density function of the standard Gaussian.
$P$ maps the posterior to a uniform distribution and $\Phi^{-1}$
maps the uniform distribution to the standard Gaussian.
The optimal choice of transformation enables the
Gaussian variational approximation to be exact. Sadly, estimating the optimal
transformation requires estimating the cumulative density function of the
posterior $P(\mbtheta \mid \mbx)$; this is just as hard as the original goal of
estimating the posterior density $p(\mbtheta \mid \mbx)$.

This observation motivates pairing transformations with Gaussian
variational approximations; there is no need for more complex
variational families. \gls{ADVI} takes the approach of using a library and
a model compiler. This is not the only option. For example,
\citet{knowles2015stochastic} posits a factorized Gamma density for positively
constrained latent variables. In theory, this is equivalent to a mean-field
Gaussian density paired with the transformation $T = P_\text{Gamma}$, the
cumulative density function of the Gamma. (In practice, $P_\text{Gamma}$ is
difficult to compute.) \citet{challis2012affine} study Fourier transform
techniques for location-scale variational approximations beyond the Gaussian.
Another option is to learn the transformation during optimization.
We discuss recent approaches in this direction in Section\nobreakspace \ref {sec:discussion}.

\section{Automatic Differentiation Variational Inference in Practice}
\label{sec:practice}

\begin{table}[tbp]
\centering
\begin{tabular}{cccc}
  \toprule
  & $\text{Gamma}(1,2)$ & $\text{Gamma}(2.5,4.2)$ & $\text{Gamma}(10,10)$\\
  \midrule
  \KL{q}{p} with $T_1$ & $8.1\times10^{-2}$ & $3.3\times10^{-2}$ & $8.5\times10^
  {-3}$\\
  \KL{q}{p} with $T_2$ & $1.6\times10^{-2}$ & $3.6\times10^{-3}$ & $7.7\times10^
  {-4}$\\
  \bottomrule
  \end{tabular}
  \caption{\gls{KL} divergence of \gls{ADVI} approximations to Gamma
densities under two different transformations.}
  \label{tab:transform_gamma}
\end{table}

\glsreset{ADVI}

We now apply \gls{ADVI} to an array of nonconjugate probability
models. With simulated and real data, we study linear regression with
automatic relevance determination, hierarchical logistic regression,
several variants of non-negative matrix factorization, mixture models,
and probabilistic principal component analysis.  We compare mean-field
\gls{ADVI} to two \gls{MCMC} sampling algorithms: \gls{HMC}
\citep{girolami2011riemann} and \gls{NUTS}, which is an adaptive
extension of \gls{HMC}\footnote{It is the default sampler in Stan.}
\citep{hoffman2014nuts}.
 
To place \gls{ADVI} and \gls{MCMC} on a common scale, we report
predictive likelihood on held-out data as a function of
time. Specifically, we estimate the predictive likelihood
\begin{align*}
  p(\mbx_{\text{held-out}} \mid \mbx)
  &=
  \int p(\mbx_{\text{held-out}} \mid \mbtheta) 
  p(\mbtheta\mid\mbx) \dif\mbtheta
\end{align*}
using Monte Carlo estimation. With \gls{MCMC}, we run the chain 
and plug in each sample to estimate the integral above; with \gls {ADVI}, we
draw a sample from the variational approximation at every iteration.

We conclude with a case study: an exploratory analysis of millions of
taxi rides. Here we show how a scientist might use \gls{ADVI} in
practice.

\subsection{Hierarchical Regression Models}
\label{sub:hierarchical_regression_models}

We begin with two nonconjugate regression models: linear regression with
\gls{ARD} \citep{bishop2006pattern} and hierarchical logistic regression
\citep{gelman2006data}.

\parhead{Linear regression with \gls{ARD}.} This is a linear
regression model with a hierarchical prior structure that leads to
sparse estimates of the coefficients.  (Details in
Appendix\nobreakspace \ref {app:linreg_ard}.)  We simulate a dataset with $250$ regressors
such that half of the regressors have no predictive power. We use
$10\,000$ data points for training and withhold $1000$ for evaluation.

\parhead{Logistic regression with a spatial hierarchical prior.} This
is a hierarchical logistic regression model from political
science. The prior captures dependencies, such as states and regions,
in a polling dataset from the United States 1988 presidential election
\citep{gelman2006data}.  The model is nonconjugate and would require
some form of approximation to derive a classical \gls{VI}
algorithm. (Details in Appendix\nobreakspace \ref {app:logreg}.)

The dataset includes $145$ regressors, with age, education, and
state and region indicators.
We use $10\,000$ data points for training and withhold $1536$ for evaluation.

\begin{figure}[!htb]
\hspace*{-0.1in}
\includegraphics{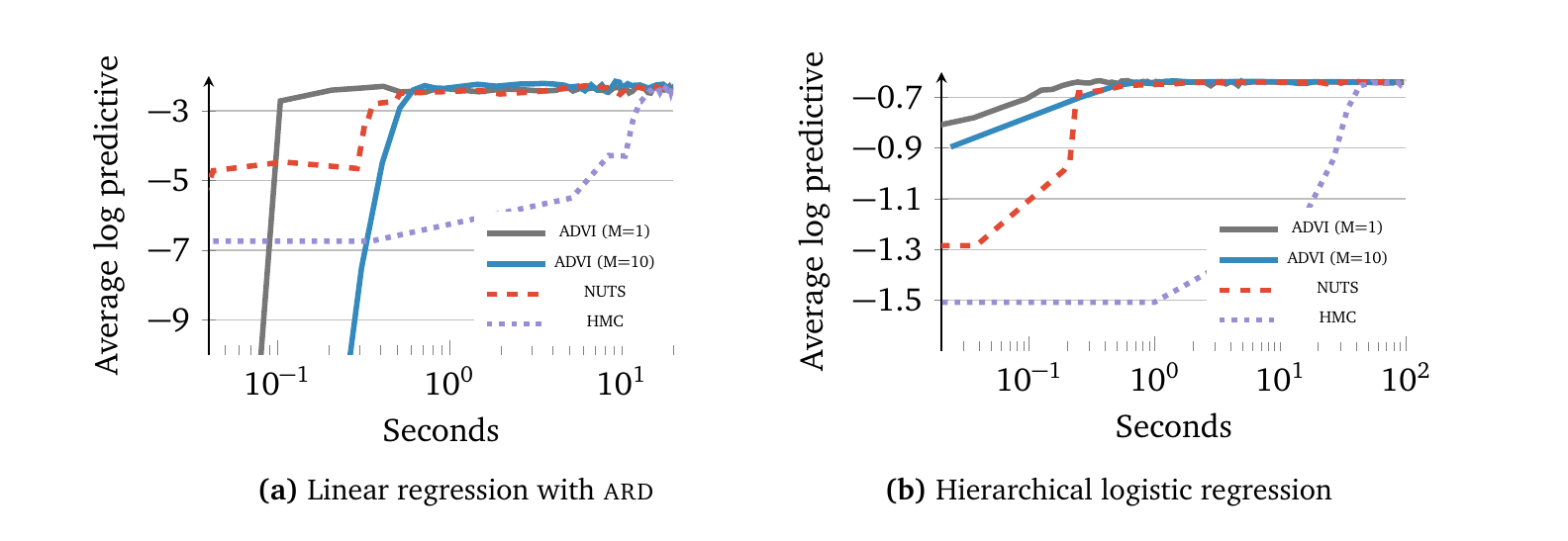}
  \caption{Held-out predictive accuracy results | hierarchical generalized
  linear models on simulated and real data.}
  \label{fig:regression_models}
\end{figure}

\parhead{Results.}
Figure\nobreakspace \ref {fig:regression_models} plots average log predictive accuracy as a
function of time. For these simple models, all methods reach the same predictive
accuracy. We study \gls{ADVI} with two settings of $M$, the number of \gls{MC}
samples used to estimate gradients. A single sample per iteration is
sufficient; it is also the fastest. (We set $M=1$ from here on.)

\subsection{Non-negative Matrix Factorization}
\label{sub:NMF}

We continue by exploring two nonconjugate non-negative matrix
factorization models~\citep{Lee:1999}: a constrained Gamma Poisson
model \citep{canny2004gap} and a Dirichlet Exponential Poisson model.
Here, we show how easy it is to explore new models using
\gls{ADVI}. In both models, we use the Frey Face dataset, which
contains $1956$ frames ($28\times20$ pixels) of facial expressions
extracted from a video sequence.

\parhead{Constrained Gamma Poisson.}
This is a Gamma Poisson matrix factorization model with an ordering
constraint: each row of one of the Gamma factors goes from small to
large values. (Details in Appendix\nobreakspace \ref {app:gap}.)

\parhead{Dirichlet Exponential Poisson.}
This is a nonconjugate Dirichlet Exponential factorization model with a Poisson
likelihood. (Details in Appendix\nobreakspace \ref {app:dir_exp}.)

\begin{figure}[!htb]
\hspace*{-0.1in}
\includegraphics{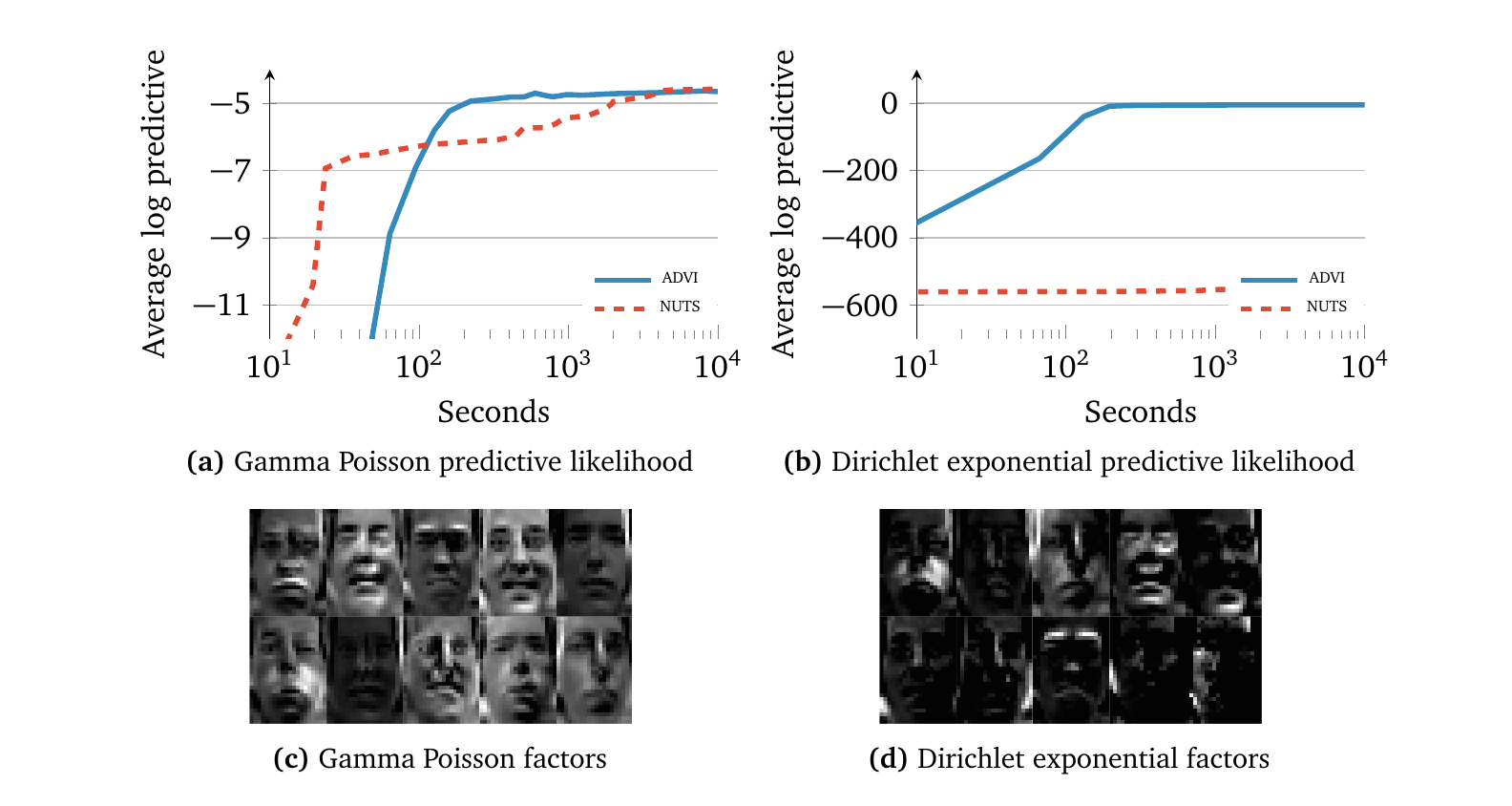}
  \caption{Held-out predictive accuracy results | two non-negative matrix
  factorization models applied to the Frey Faces dataset.}
  \label{fig:nmf}
\end{figure}

\parhead{Results.}
Figure\nobreakspace \ref {fig:nmf} shows average log predictive accuracy as well as ten factors
recovered from both models.
\gls{ADVI} provides an order of magnitude speed improvement over \gls{NUTS}. 
\gls{NUTS} struggles with the Dirichlet Exponential
model. In both cases, \gls{HMC} does not
produce any useful samples within a budget of one hour; we omit \gls{HMC} from
here on.

The Gamma Poisson model appears to pick
significant frames out of the dataset. The Dirichlet Exponential factors
are sparse and indicate
components of the face that move, such as eyebrows, cheeks, and the mouth.

\subsection{Gaussian Mixture Model}
\label{sub:gmm}

This is a nonconjugate \gls {GMM} applied to color image
histograms. We place a Dirichlet prior on the mixture proportions, a
Gaussian prior on the component means, and a lognormal prior on the
standard deviations. (Details in Appendix\nobreakspace \ref {app:gmm}.) We explore the
image\textsc{clef} dataset, which has $250\,000$ images
\citep{villegas13_CLEF}. We withhold $10\,000$ images for evaluation.

\begin{figure}[htb]
\hspace*{-0.1in}
\includegraphics{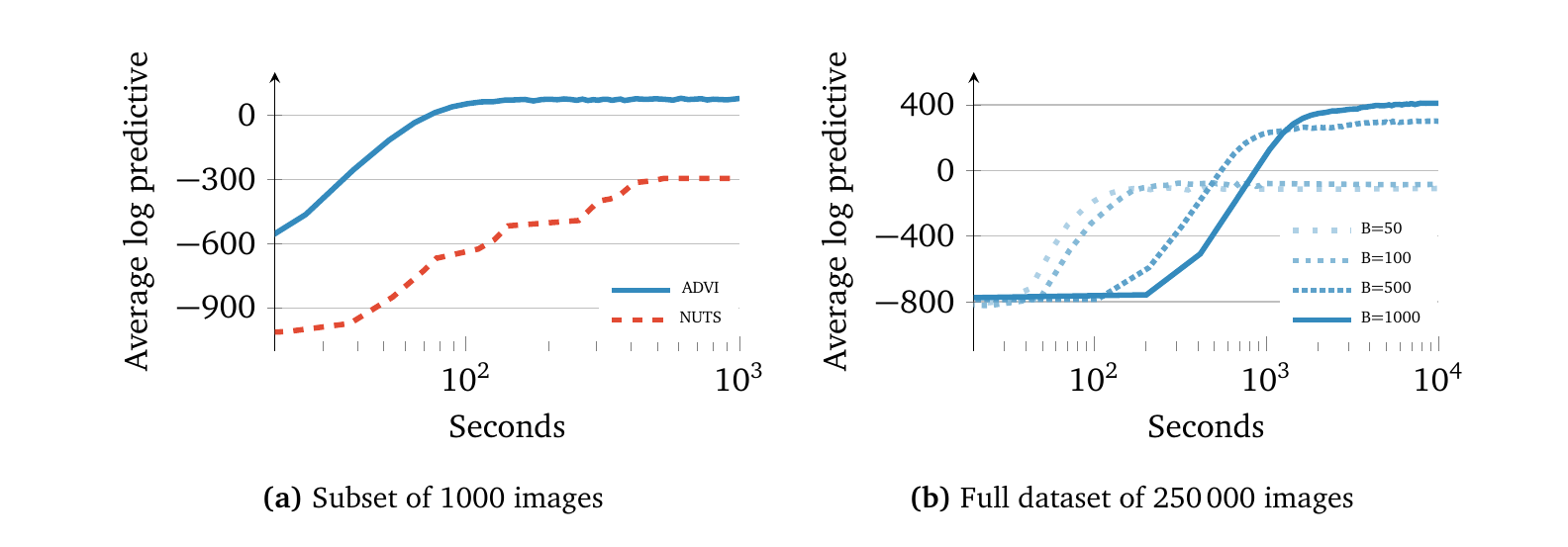}
  \caption{
  Held-out predictive accuracy results | \gls{GMM} of the image\textsc
  {clef} image histogram dataset. \textbf{(a)} \gls{ADVI}
  outperforms \gls{NUTS} \citep{hoffman2014nuts}.
  \textbf{(b)} \gls{ADVI} scales to large datasets by subsampling minibatches
  of size $B$ from the dataset at each iteration \citep{hoffman2013stochastic}.
  }
  \label{fig:gmm_plots}
\end{figure}

In Figure\nobreakspace \ref {fig:gmm_plots}a we randomly select $1000$ images and train a
model with $10$ mixture components. \gls{ADVI} quickly finds a good
solution.  \gls{NUTS} struggles to find an adequate solution and
\gls{HMC} fails altogether (not shown). This is likely due to label switching,
which can affect \gls{HMC}-based algorithms in mixture models
\citep{stan-manual:2015}.

Figure\nobreakspace \ref {fig:gmm_plots}b shows \gls{ADVI} results on the full dataset.  We
increase the number of mixture components to $30$. Here we use
\gls{ADVI}, with additional stochastic subsampling of
minibatches from the data \citep{hoffman2013stochastic}. With a
minibatch size of $500$ or larger, \gls{ADVI} reaches high predictive
accuracy. Smaller minibatch sizes lead to suboptimal solutions, an
effect also observed in \citet{hoffman2013stochastic}. \gls{ADVI}
converges in about two hours; \gls{NUTS} cannot handle such large
datasets.

\glsreset{ARD}
\subsection{A Case Study: Exploring Millions of Taxi Trajectories}
\label{sub:taxicab}

How might a scientist use \gls{ADVI} in practice? How easy is it to
develop and revise new models? To answer these questions, we apply \gls{ADVI} to
a modern exploratory data analysis task: analyzing traffic patterns.
In this section, we demonstrate how \gls{ADVI} enables
a scientist to quickly develop and revise complex hierarchical
models.

The city of Porto has a centralized taxi system of 442 cars.
When serving customers, each taxi reports its spatial location at 15
second intervals; this sequence of $(x,y)$ coordinates describes the trajectory
and duration of each trip. A dataset of trajectories is publicly
available: it contains all 1.7 million taxi rides taken during the year 2014
(European Conference of Machine Learning, 2015).

To gain insight into this dataset, we wish to cluster the
trajectories. The first task is to process the raw data. Each
trajectory has a different length: shorter trips contain fewer $(x,y)$
coordinates than longer ones.  The average trip is approximately 13
minutes long, which corresponds to 50 coordinates.  We want to cluster
independent of length, so we interpolate all trajectories to 50
coordinate pairs.  This converts each trajectory into a point in
$\bbR^{100}$.

The trajectories have structure; for example, major roads and highways
appear frequently.  This motivates an approach where we first identify
a lower-dimensional representation of the data to capture aggregate
features, and then we cluster the trajectories in this representation.
This is easier than clustering them in the original data space.

We begin with simple dimension reduction: \gls{PPCA}
\citep{bishop2006pattern}. This is a Bayesian generalization of
classical principal component analysis, which is easy to write in
Stan.  However, like its classical counterpart, \gls{PPCA} does not
identify how many principal components to use for the subspace. To
address this, we propose an extension: \gls{PPCA} with \gls{ARD}.

\gls{PPCA} with \gls{ARD} identifies the latent dimensions that are
most effective at explaining variation in the data. The strategy is
similar to that in Section\nobreakspace \ref {sub:hierarchical_regression_models}.  We
assume that there are $100$ latent dimensions (i.e., the same
dimension as the data) and impose a hierarchical prior that encourages
sparsity.  Consequently, the model only uses a subset of the latent
dimensions to describe the data. (Details in Appendix\nobreakspace \ref {app:ppca_ard2}.)

\begin{figure}[!htb]
\hspace*{-0.1in}
\includegraphics{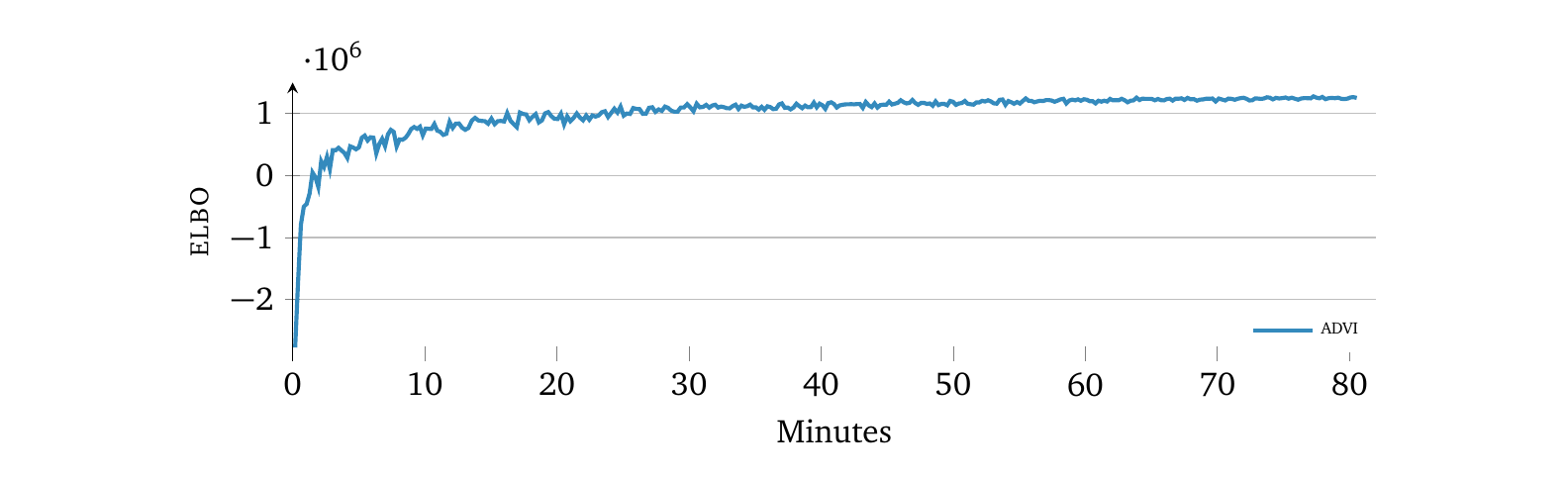}
  \caption{\gls{ELBO} of \gls{PPCA} model with \gls{ARD}. \gls{ADVI} converges
  in approximately an hour.}
  \label{fig:ppca_ard2}
\end{figure}

We randomly subsample ten thousand trajectories and use \gls{ADVI} to
infer a subspace. Figure\nobreakspace \ref {fig:ppca_ard2} plots the progression of the
\gls{ELBO}. \gls {ADVI} converges in approximately an hour and finds
an eleven-dimensional subspace. We omit sampling results as both
\gls{HMC} and \gls{NUTS} struggle with the model; neither produce
useful samples within an hour.

Equipped with this eleven-dimensional subspace, we turn to analyzing
the full dataset of 1.7 million taxi trajectories. We first project
all trajectories into the subspace. We then use the \gls{GMM} from
Section\nobreakspace \ref {sub:gmm} ($K=30$) components to cluster the
trajectories. \gls{ADVI} takes less than half an hour to converge.

\begin{figure}[!htb]
\centering
  \includegraphics[width=3in]{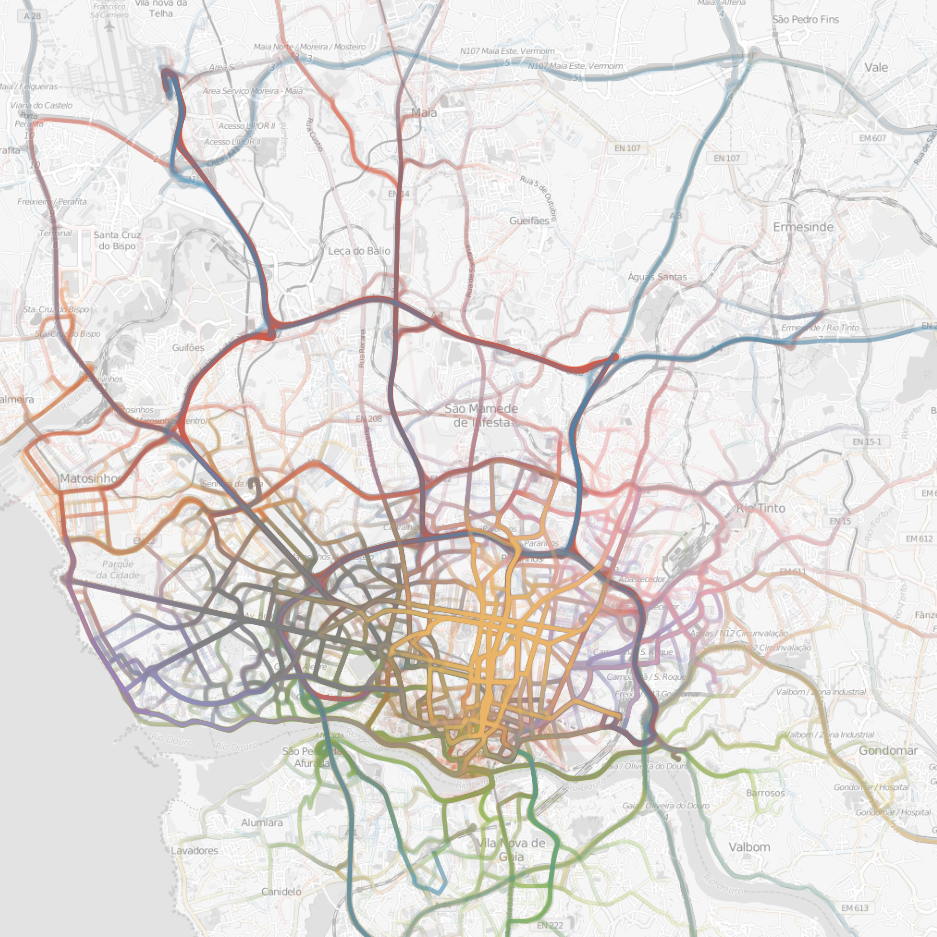}
  \caption{A visualization of fifty thousand randomly sampled taxi
  trajectories. The colors represent thirty Gaussian mixtures and the
  trajectories associated with each.} \label {fig:porto_ppca}
\end{figure}

Figure\nobreakspace \ref {fig:porto_ppca} shows a visualization of fifty thousand randomly
sampled trajectories. Each color represents the set of trajectories
that associate with a particular Gaussian mixture. The clustering is
geographical: taxi trajectories that are close to each other are
bundled together. The clusters identify frequently taken taxi
trajectories.

When we processed the raw data, we interpolated each trajectory to an
equal length.  This discards all duration information. What if some
roads are particularly prone to traffic? Do these roads lead to longer
trips?

\Gls{SUP-PPCA} is one way to model this. The idea is to regress the
durations of each trip onto a subspace that also explains variation in
a response variable, in this case, the duration. \gls{SUP-PPCA} is a
simple extension of \gls{PPCA} \citep{murphy2012machine}. We further
extend it using the same \gls{ARD} prior as before. 
(Details in Appendix\nobreakspace \ref {app:sup_ppca_ard2}.)

\gls{ADVI} enables a quick repeat of the above analysis, this time
with \gls {SUP-PPCA}. With \gls{ADVI}, we find another set of
\gls{GMM} clusters in less than two hours. These clusters, however,
are more informative.

\begin{figure}[!htb]
\centering
  \begin{subfigure}[b]{2.6in}
  \centering
    \includegraphics[trim={0 150px 50px 300px},clip,width=2.5in]
    {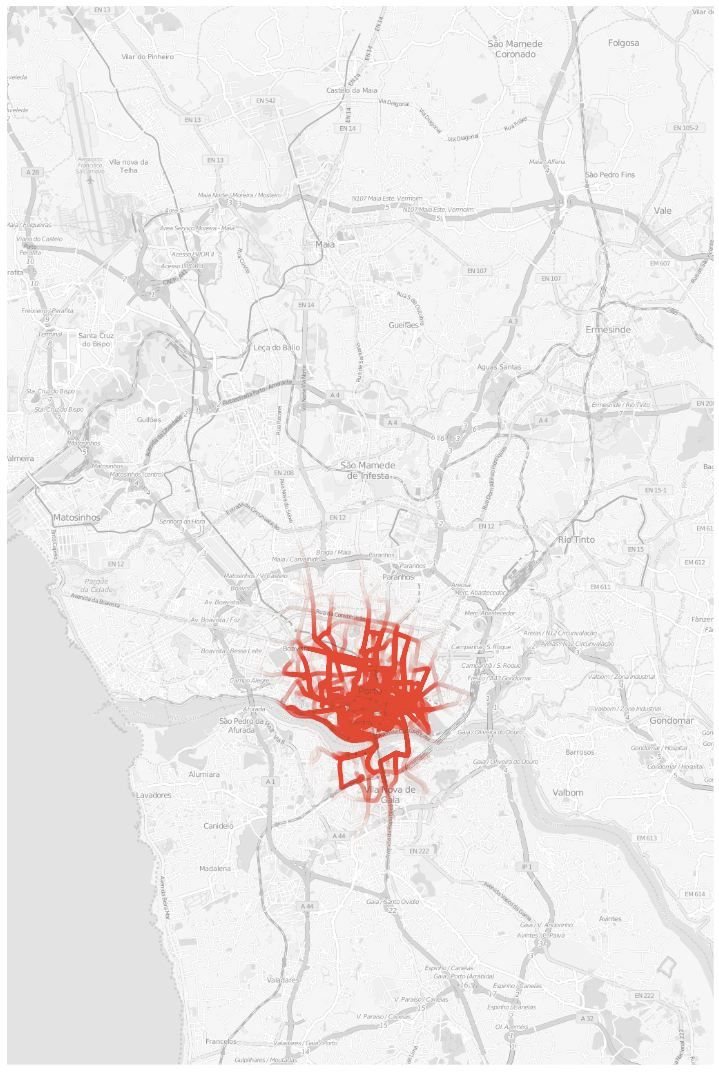}
    \caption{Trajectories that take the inner bridges.}
    \label{sub:sup_ppca_ard2_26}
  \end{subfigure}
  \begin{subfigure}[b]{2.6in}
  \centering
    \includegraphics[trim={0 150px 50px 300px},clip,width=2.5in]
    {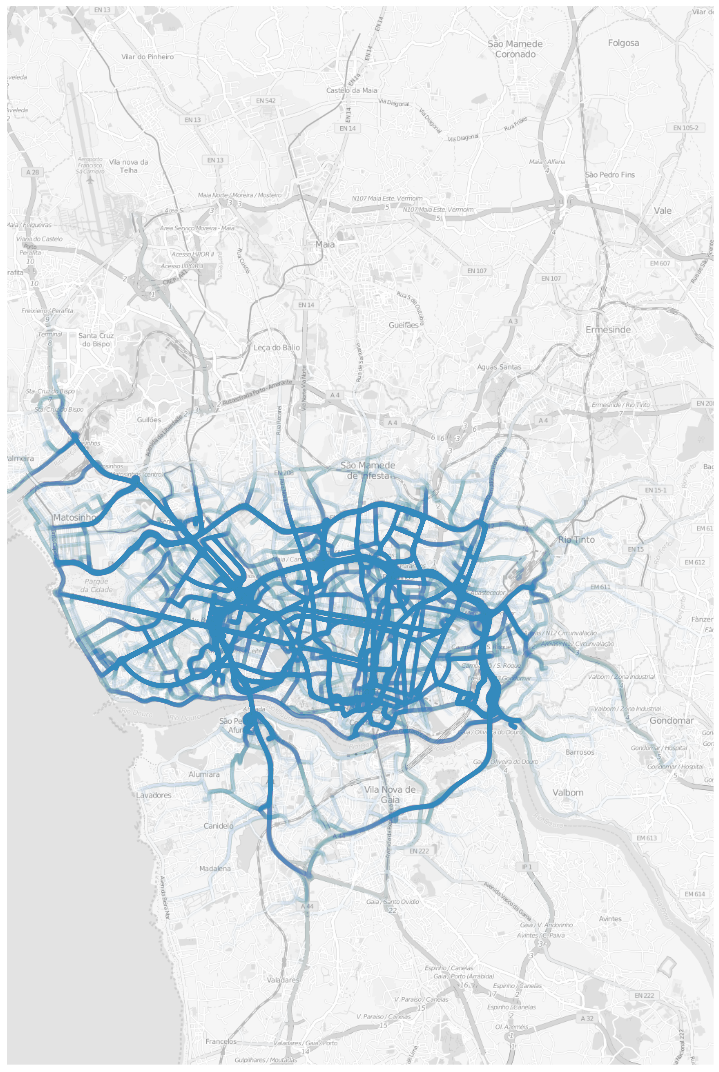}
    \caption{Trajectories that take the outer bridges.}
    \label{sub:sup_ppca_ard2_18}
  \end{subfigure}
  \caption{Two clusters using \gls{SUP-PPCA} subspace clustering.}
  \label{fig:sup_ppca_ard2}
\end{figure}

Figure\nobreakspace \ref {fig:sup_ppca_ard2} shows two clusters that identify particularly
busy roads: the bridges of Porto that cross the Duoro river.
Figure\nobreakspace \ref {sub:sup_ppca_ard2_26} shows a group of short trajectories that
use the two old bridges near the city
center. Figure\nobreakspace \ref {sub:sup_ppca_ard2_18} show a group of longer
trajectories that use the two newer bridges that connect highways that
circumscribe the city.

Analyzing these taxi trajectories illustrates how exploratory data
analysis is an iterative effort: we want to rapidly evaluate models
and modify them based on what we learn.  \gls{ADVI}, which provides
automatic and fast inference, enables effective exploration of massive
datasets.

\glsreset{ADVI}
\section{Discussion}
\label{sec:discussion}

We presented \gls{ADVI}, a variational inference tool that works for a
large class of probabilistic models. The main idea is to transform the
latent variables into a common space. Solving the variational
inference problem in this common space solves it for all models in the
class. We studied \gls{ADVI} using ten different probability models;
this showcases how easy it is to use \gls{ADVI} in practice. We also
developed and deployed \gls{ADVI} as part of Stan, a probabilistic
programming system; this makes \gls{ADVI} available to everyone.

There are several avenues for research.

Begin with accuracy. As we showed in
Section\nobreakspace \ref {sub:sensitivity_to_transformations}, \gls{ADVI} can be sensitive
to the transformations that map the constrained parameter space to the
real coordinate space.  \citet{dinh2014nice} and
\citet{rezende2015variational} use a cascade of simple transformations
to improve accuracy.  \citet{tran2015variational} place a Gaussian
process to learn the optimal transformation and prove its
expressiveness as a universal approximator.  A class of hierarchical
variational models \citep{ranganath2015hierarchical} extend these
complex distributions to discrete latent variable models.

Continue with optimization. \gls{ADVI} uses first-order automatic
differentiation to implement stochastic gradient ascent. Higher-order
gradients may enable faster convergence; however computing higher-order
gradients comes at a computational cost \citep{fan2015fast}. Optimization using
line search could also improve convergence speed and robustness
\citep{mahsereci2015}, as well as natural gradient approaches for
nonconjugate models \citep{khan2015kullback}.

Follow with practical heuristics. Two things affect \gls{ADVI}
convergence: initialization and step-size scaling. We initialize
\gls{ADVI} in the real coordinate space as a standard
Gaussian. A better heuristic could adapt to the model and dataset based
on moment matching. We
adaptively tune the scale of the step-size sequence using a finite
search. A better heuristic could avoid this additional computation.

End with probabilistic programming. We designed and deployed
\gls{ADVI} with Stan in mind. Thus, we focused on the class of
differentiable probability models. How can we extend \gls{ADVI} to
discrete latent variables? One approach would be to adapt \gls{ADVI} to use the
black box gradient estimator for these variables
\citep{ranganath2014black}. This requires some care as these gradients
will exhibit higher variance than the gradients with respect to
the differentiable latent variables.  (See
Section\nobreakspace \ref {sub:variance_of_gradient_estimators}.) With support for discrete
latent variables, modified versions of \gls{ADVI} could be extended to
more general probabilistic programming systems, such as Church
\citep{goodman2008church}, Figaro \citep{pfeffer2009figaro}, Venture
\citep{mansinghka2014venture}, and Anglican \citep{wood2014new}.

\parhead{Acknowledgments.}
We thank Bruno Jacobs, and the reviewers for their helpful comments. 
This work is supported by NSF IIS-0745520, IIS-1247664, IIS-1009542,
SES-1424962, ONR N00014-11-1-0651, DARPA FA8750-14-2-0009, N66001-15-C-4032,
Sloan G-2015-13987, IES DE R305D140059, NDSEG, Facebook, Adobe, Amazon, and the
Siebel Scholar and John Templeton Foundations.

\clearpage
\appendix

\section{Transformations of Continuous Probability Densities}
\label{app:jacobian}

We present a brief summary of transformations, largely based on \citep
{olive2014statistical}.

Consider a scalar (univariate) random variable $X$ with probability density
function $f_X(x)$. Let $\mathcal{X} = \supp(f_X(x))$ be the support of
$X$. Now consider another random variable $Y$ defined as $Y = T(X)$.
Let $\mathcal{Y} = \supp(f_Y(y))$ be the support of $Y$.

If $T$ is a one-to-one and differentiable function from $\mathcal{X}$ to
$\mathcal{Y}$, then $Y$ has probability density function
\begin{align*}
  f_Y(y)
  &=
  f_X\left(T^{-1}(y)\right)
  \left|
  \frac{\dif T^{-1}(y)}{\dif y}
  \right|.
\end{align*}
Let us sketch a proof. Consider the cumulative density function
$Y$. If the transformation $T$ is increasing, we directly apply its
inverse to the cdf of $Y$. If the transformation $T$ is decreasing, we apply its
inverse to one minus the cdf of $Y$. The probability density function is the
derivative of the cumulative density function. These things combined give the
absolute value of the derivative above.

The extension to multivariate variables $\mb{X}$ and $\mb{Y}$ requires a
multivariate version of the absolute value of the derivative of the inverse
transformation. This is the absolute determinant of the Jacobian,
$|\det J_{T^{-1}}(\mb{Y})|$ where the Jacobian is
\begin{align*}
  J_{T^{-1}}(\mb{Y})
  &=
  \left(
  \begin{matrix}
    \frac{\partial T_1^{-1}}{\partial y_1}
    &
    \cdots
    &
    \frac{\partial T_1^{-1}}{\partial y_K}\\
    \vdots & & \vdots\\
    \frac{\partial T_K^{-1}}{\partial y_1}
    &
    \cdots
    &
    \frac{\partial T_K^{-1}}{\partial y_K}\\
  \end{matrix}
  \right).
\end{align*}

Intuitively, the Jacobian describes how a transformation warps unit volumes
across spaces. This matters for transformations of random variables, since
probability density functions must always integrate to one. If the
transformation is linear, then we can drop the Jacobian adjustment; it evaluates
to one. Similarly, affine transformations, like elliptical standardizations,
also have Jacobians that evaluate to one; they preserve unit volumes.

\section{Transformation of the Evidence Lower Bound}
\label{app:elbo_unconstrained}

Recall that $\mbzeta = T(\mbtheta)$ and that the variational approximation in
the real coordinate space is
$q(\mbzeta \,;\, \mbphi)$.

We begin with the \gls{ELBO} in the original latent variable space. We then
transform the latent variable space to the real coordinate space.
\begin{align*}
  \cL(\mbphi)
  &=
  \int
  q(\mbtheta)
  \log \left[\frac{p(\mbx,\mbtheta)}{q(\mbtheta)}\right]
  \dif\mbtheta
  \\
  &=
  \int
  q(\mbzeta \,;\, \mbphi)
  \log
  \left[
  \frac
  {p \left(\mbx,T^{-1}(\mbzeta)\right) \big| \det J_{T^{-1}}(\mbzeta) \big|}
  {q(\mbzeta \,;\, \mbphi)}
  \right]
  \dif \mbzeta
  \\
  &=
  \int q(\mbzeta \,;\, \mbphi)
  \log
  \left[
  p \left(\mbx,T^{-1}(\mbzeta)\right) \big| \det J_{T^{-1}}(\mbzeta) \big|
  \right]
  \dif \mbzeta
  -
  \int q(\mbzeta \,;\, \mbphi)
  \log
  \left[
  q(\mbzeta \,;\, \mbphi)
  \right]
  \dif \mbzeta
  \\
  &= \E_{q(\mbzeta\,;\, \mbphi)} \left[ \log p \left(\mbx,T^{-1}(\mbzeta)\right)
  +
  \log \big| \det J_{T^{-1}}(\mbzeta) \big| \right]
  -
  \E_{q(\mbzeta\,;\, \mbphi)} \left[ \log q(\mbzeta \,;\, \mbphi) \right]
  \\
  &=
  \E_{q(\mbzeta\,;\, \mbphi) } \left[ \log p \left(\mbx,T^{-1}(\mbzeta)\right)
  +
  \log \big| \det J_{T^{-1}}(\mbzeta) \big| \right]
  +
  \bbH
  \big[
  q(\mbzeta\,;\,\mbphi)
  \big].
\end{align*}

\section{Gradients of the Evidence Lower Bound}
\label{app:gradient_elbo}

First, consider the gradient with respect to the $\mbmu$ parameter. 
We exchange the order of the gradient and the integration
through the dominated convergence theorem \citep{cinlar2011probability}. The
rest is the chain rule for differentiation.
\begin{align*}
  \nabla_\mbmu \cL
  &=
  \nabla_\mbmu
  \Big\{
  \E_{\cN(\mbeta\,;\, \mb{0}, \mbI)}
  \left[
  \log p
  \left(\mbx,T^{-1}(S_{\mbphi}^{-1}(\mbeta))\right)
  +
  \log \big| \det J_{T^{-1}}\left(S_{\mbphi}^{-1}(\mbeta)\right) \big|
  \right]
  +
  \bbH
  \big[
  q(\mbzeta\,;\,\mbphi)
  \big]
  \Big\}\\
  &=
  \E_{\cN(\mbeta\,;\, \mb{0}, \mbI)}
  \left[
  \nabla_\mbmu
  \left\{
  \log p
  \left(\mbx,T^{-1}(S^{-1}(\mbeta))\right)
  +
  \log \big| \det J_{T^{-1}}\left(S^{-1}(\mbeta)\right) \big|
  \right\}
  \right] \\
  &=
  \E_{\cN(\mbeta\,;\, \mb{0}, \mbI)}
  \left[
  \left(
  \nabla_\mbtheta \log p(\mbx,\mbtheta)
  \nabla_{\mbzeta} T^{-1}(\mbzeta)
  +
  \nabla_{\mbzeta}
  \log \big| \det J_{T^{-1}}(\mbzeta) \big|
  \right)
  \nabla_{\mbmu} S_{\mbphi}^{-1} (\mbeta)
  \right] \\
  &=
  \E_{\cN(\mbeta\,;\, \mb{0}, \mbI)}
  \left[
  \nabla_\mbtheta \log p(\mbx,\mbtheta)
  \nabla_{\mbzeta} T^{-1}(\mbzeta)
  +
  \nabla_{\mbzeta}
  \log \big| \det J_{T^{-1}}(\mbzeta) \big|
  \right]
\end{align*}

Then, consider the gradient with respect to the mean-field $\mbomega$
parameter.
\begin{align*}
  \nabla_{\mbomega} \cL
  &=
  \nabla_{\mbomega}
  \Big\{
  \E_{\cN(\mbeta\,;\, \mb{0}, \mbI)}
  \left[
  \log p
  \left(\mbx,T^{-1}(S_{\mbphi}^{-1}(\mbeta))\right)
  +
  \log \big| \det J_{T^{-1}}\left(S_{\mbphi}^{-1}(\mbeta)\right) \big|
  \right]\\
  &\qquad\quad+ \frac{K}{2}(1+\log(2\pi))
  + \sum_{k=1}^K \log(\exp(\omega_k))
  \Big\}\\
  &=
  \E_{\cN(\mbeta\,;\, \mb{0}, \mbI)}
  \left[
  \nabla_{\mbomega}
  \big\{
  \log p
  \left(\mbx,T^{-1}(S_{\mbphi}^{-1}(\mbeta))\right)
  +
  \log \big| \det J_{T^{-1}}\left(S_{\mbphi}^{-1}(\mbeta)\right) \big|
  \big\}
  \right] + \mbone\\
  &=
  \E_{\cN(\mbeta\,;\, \mb{0}, \mbI)}
  \left[
  \left(
  \nabla_{\mbtheta} \log p(\mbx,\mbtheta)
  \nabla_{\mbzeta} T^{-1}(\mbzeta)
  +
  \nabla_{\mbzeta}
  \log \big| \det J_{T^{-1}}(\mbzeta) \big|
  \right)
  \nabla_{\mbomega} S_{\mbphi}^{-1}(\mbeta))
  \right]
  + \mbone\\
  &=
  \E_{\cN(\mbeta\,;\, \mb{0}, \mbI)}
  \left[
  \left(
  \nabla_{\mbtheta} \log p(\mbx,\mbtheta)
  \nabla_{\mbzeta} T^{-1}(\mbzeta)
  +
  \nabla_{\mbzeta}
  \log \big| \det J_{T^{-1}}(\mbzeta) \big|
  \right)
  \mbeta^\top
  \diag(\exp(\mbomega))
  \right]
  + \mbone.
\end{align*}

Finally, consider the gradient with respect to the full-rank $\mbL$
parameter.
\begin{align*}
  \nabla_{\mbL}
  \cL
  &=
  \nabla_{\mbL}
  \Big\{
  \E_{\cN(\mbeta\,;\, \mb{0}, \mbI)}
  \left[
  \log p
  \left(\mbx,T^{-1}(S_{\mbphi}^{-1}(\mbeta))\right)
  +
  \log \big| \det J_{T^{-1}}\left(S_{\mbphi}^{-1}(\mbeta)\right) \big|
  \right]\\
  &\qquad\quad+ \frac{K}{2}(1+\log(2\pi))
  + \frac{1}{2} \log \big| \det (\mbL\mbL^\top) \big|
  \Big\}\\
  &=
  \E_{\cN(\mbeta\,;\, \mb{0}, \mbI)}
  \left[
  \nabla_{\mbL}
  \big\{
  \log p
  \left(\mbx,T^{-1}(S_{\mbphi}^{-1}(\mbeta))\right)
  +
  \log \big| \det J_{T^{-1}}\left(S_{\mbphi}^{-1}(\mbeta)\right) \big|
  \big\}
  \right]\\
  &\qquad\quad+ \nabla_{\mbL}\frac{1}{2} \log \big| \det (\mbL\mbL^\top) \big|\\
  &=
  \E_{\cN(\mbeta\,;\, \mb{0}, \mbI)}
  \left[
  \left(
  \nabla_\mbtheta \log p(\mbx,\mbtheta)
  \nabla_{\mbzeta} T^{-1}(\mbzeta)
  +
  \nabla_{\mbzeta}
  \log \big| \det J_{T^{-1}}(\mbzeta) \big|
  \right)
  \nabla_\mbL S_{\mbphi}^{-1}(\mbeta))
  \right] + (\mbL^{-1})^\top\\
  &=
  \E_{\cN(\mbeta\,;\, \mb{0}, \mbI)}
  \left[
  \left(
  \nabla_\mbtheta \log p(\mbx,\mbtheta)
  \nabla_{\mbzeta} T^{-1}(\mbzeta)
  +
  \nabla_{\mbzeta}
  \log \big| \det J_{T^{-1}}(\mbzeta) \big|
  \right)
  \mbeta^\top
  \right] + (\mbL^{-1})^\top\\
\end{align*}

\section{Automating Expectations: Monte Carlo Integration}
\label{app:mc_integration}

Expectations are integrals. We can use \gls{MC} integration to approximate them
\citep{robert1999monte}. All we need are samples from $q$.
\begin{align*}
  \E_{q(\mbeta)}
  \big[
  f(\mbeta)
  \big]
  &=
  \int
  f(\mbeta)
  q(\mbeta)
  \dif \mbeta
  \approx
  \frac{1}{S}
  \sum_{s=1}^S
  f(\mbeta_s)
  \text{  where  }
  \mbeta_s \sim q(\mbeta).
\end{align*}

$\gls{MC}$ integration provides noisy, yet unbiased, estimates of the integral.
The standard deviation of the estimates are of order $1/\sqrt{S}$.

\section{Running \textsc{advi} in Stan}
\label{app:stan}

Visit \texttt{http://mc-stan.org/} to download the latest version of Stan.
Follow
instructions on how to install Stan. You are then ready to use \gls{ADVI}.

Stan offers multiple interfaces. We describe the command line
interface (\texttt{cmdStan}) below,
\begin{figure}[!h]
\centering
\ttfamily
\begin{tabular}{rl}
./myModel & variational\\
          & grad\_samples=M $\qquad\qquad\qquad$( $M = 1$ default )\\
          & data file=myData.data.R\\
          & output file=output\_advi.csv\\
          & diagnostic\_file=elbo\_advi.csv
\end{tabular}
\caption{Syntax for using \gls{ADVI} via \texttt{cmdStan}.}
\end{figure}
where \texttt{myData.data.R} is the dataset stored in the \texttt{R} language
\texttt{Rdump} format. \texttt{output\_advi.csv} contains samples from
the posterior and \texttt{elbo\_advi.csv} reports the \gls{ELBO}.

\section{Details of Studied Models}

\subsection{Linear Regression with Automatic Relevance Determination}
\label{app:linreg_ard}

\begin{figure}[htbp]
\includegraphics{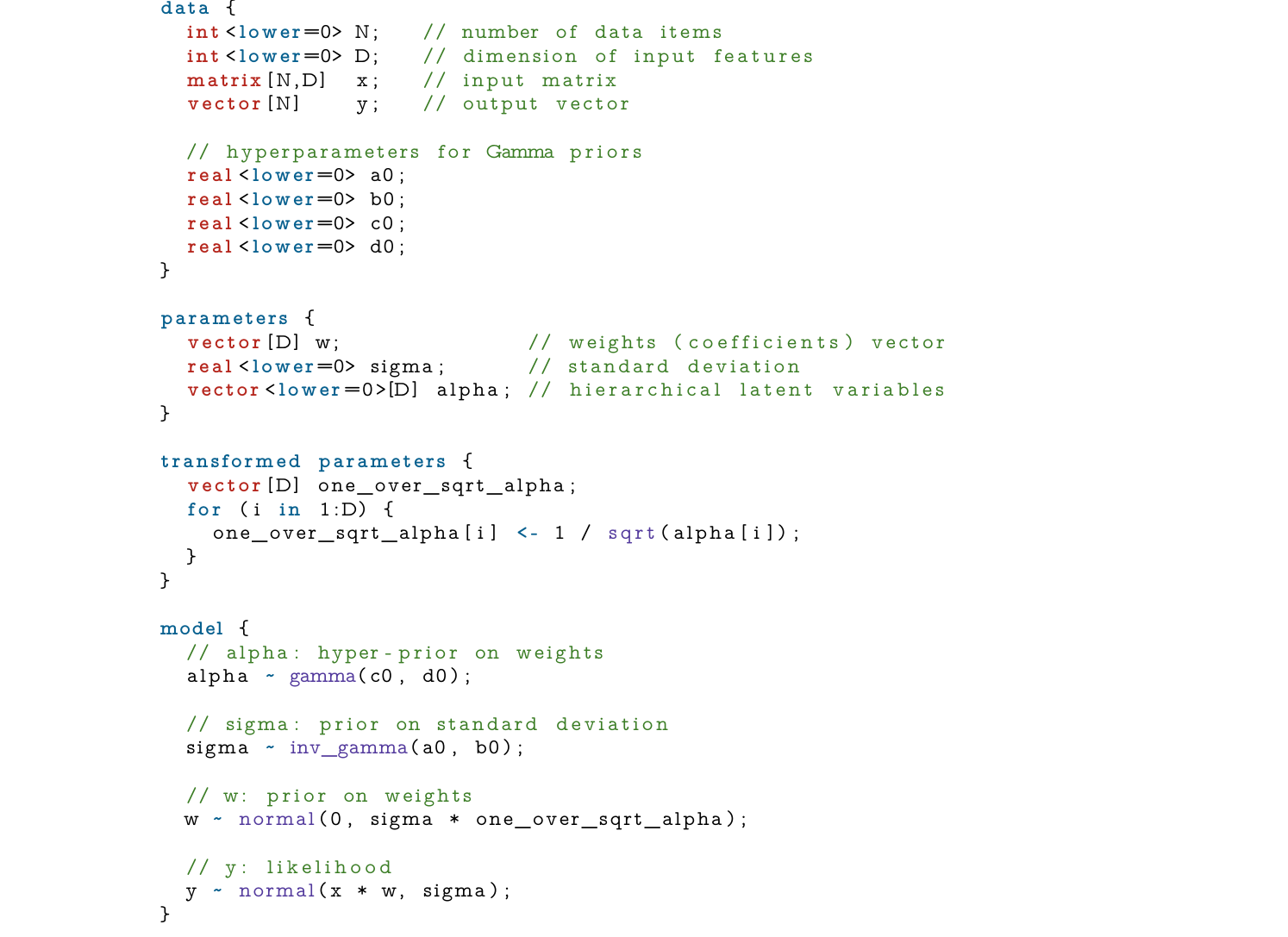}
  \caption{Stan code for Linear Regression with Automatic Relevance
  Determination.}
  \label{fig:code_linreg}
\end{figure}

Linear regression with \gls{ARD} is a high-dimensional sparse
regression model \citep{bishop2006pattern,drugowitsch2013variational}. We
describe the model below. Stan code is in Figure\nobreakspace \ref {fig:code_linreg}.

The inputs are $\mbx=x_{1:N}$ where each $x_n$ is $D$-dimensional. The
outputs are $\mb{y}=y_{1:N}$ where each $y_n$ is $1$-dimensional. The weights
vector $\mb{w}$ is
$D$-dimensional. The likelihood
\begin{align*}
  p(\mb{y} \mid \mbx, \mb{w}, \sigma)
  &=
  \prod_{n=1}^N \cN \left( y_n \mid \mb{w}^\top \mbx_n \;,\; \sigma\right)
\end{align*}
describes measurements corrupted by iid Gaussian noise with unknown
standard deviation $\sigma$.

The \gls{ARD} prior and hyper-prior structure is as follows
\begin{align*}
  p(\mb{w},\sigma,\mb{\alpha})
  &=
  p(\mb{w},\sigma \mid \mb{\alpha})
  p(\mb{\alpha})\\
  &=
  \cN \left( \mb{w} \mid 0 \,,\, \sigma \left(\diag \mb{\sqrt\alpha}\right)^{-1}
  \right)
  \InvGam (\sigma \mid a_0, b_0)
  \prod_{i=1}^D \Gam({\alpha}_i \mid c_0, d_0)
\end{align*}
where $\mb{\alpha}$ is a $D$-dimensional hyper-prior on the weights, where each
component gets its own independent Gamma prior.

We simulate data such that only half the regressions have predictive power.
The results in Figure\nobreakspace \ref {fig:regression_models} use $a_0 = b_0 = c_0 = d_0 = 1$ as
hyper-parameters for the Gamma priors.

\subsection{Hierarchical Logistic Regression}
\label{app:logreg}

\begin{figure}[htbp]
\includegraphics{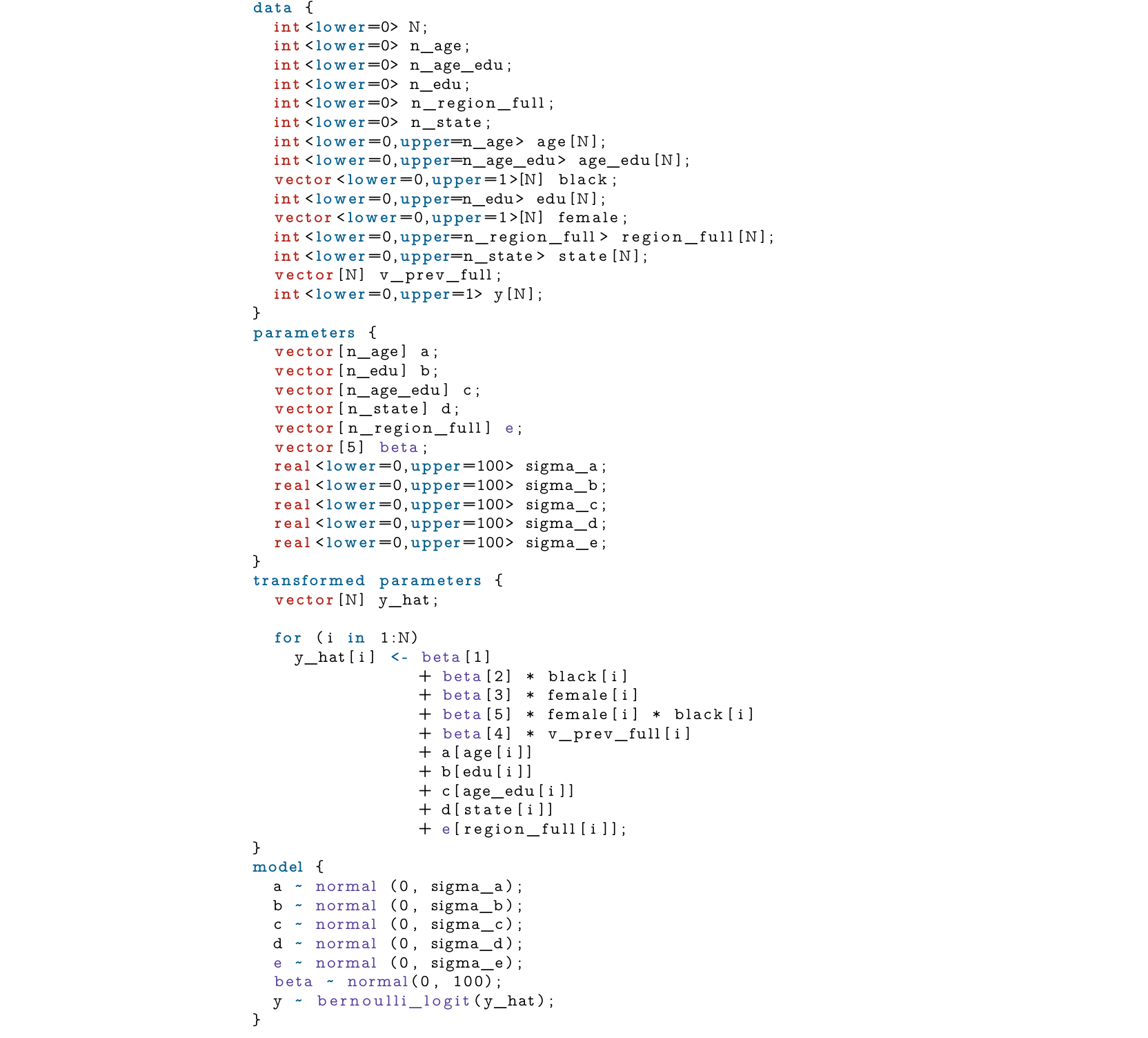}
  \caption{Stan code for Hierarchical Logistic Regression,
  from \citep{stan-manual:2015}.}
  \label{fig:code_election88}
\end{figure}

Hierarchical logistic regression models structured datasets in an intuitive way.
We study a model of voting preferences from the 1988 United
States presidential election. Chapter 14.1 of \citep{gelman2006data}
motivates the model and explains the dataset. We also describe the
model below.
Stan code is in Figure\nobreakspace \ref {fig:code_election88}, based on \citep{stan-manual:2015}.
\begin{align*}
  \Pr(y_n=1)
  &=
  \text{sigmoid}
  \bigg(
  \beta^0
  + \beta^\text{female}\cdot\text{female}_n
  + \beta^\text{black}\cdot\text{black}_n
  + \beta^\text{female.black}\cdot\text{female.black}_n \\
  &\qquad\quad + \alpha^\text{age}_{k[n]}
  + \alpha^\text{edu}_{l[n]}
  + \alpha^\text{age.edu}_{k[n],l[n]}
  + \alpha^\text{state}_{j[n]}
  \bigg)\\
  \alpha^\text{state}_j
  &\sim
  \cN
  \left(
  \alpha^\text{region}_{m[j]}
  + \beta^\text{v.prev}\cdot\text{v.prev}_j
  \,,\, \sigma_\text{state}
  \right).
\end{align*}

The hierarchical variables are
\begin{align*}
  \alpha^\text{age}_k
  &\sim
  \cN \left(0\,,\, \sigma_\text{age} \right)
  \text{ for } k = 1,\ldots,K\\
  \alpha^\text{edu}_l
  &\sim
  \cN \left(0\,,\, \sigma_\text{edu} \right)
  \text{ for } l = 1,\ldots,L\\
  \alpha^\text{age.edu}_{k,l}
  &\sim
  \cN \left(0\,,\, \sigma_\text{age.edu} \right)
  \text{ for } k = 1,\ldots,K, l = 1,\ldots,L\\
  \alpha^\text{region}_m
  &\sim
  \cN \left(0\,,\, \sigma_\text{region} \right)
  \text{ for } m = 1,\ldots,M.
\end{align*}

The standard deviation terms all have uniform hyper-priors, constrained between
0 and 100.

\subsection{Non-negative Matrix Factorization: Constrained Gamma Poisson Model}
\label{app:gap}

The Gamma Poisson factorization model describes discrete data matrices \citep
{canny2004gap,cemgil2009bayesian}.

Consider a $U \times I$ matrix of observations. We find it helpful
to think of $u=\{1,\cdots,U\}$ as users and $i=\{1,\cdots,I\}$ as items, as in a
recommendation system
setting. The generative process for a Gamma Poisson model with $K$ factors is
\begin{enumerate}
  \item For each user $u$ in $\{1,\cdots,U\}$:
  \begin{itemize}
    \item For each component $k$, draw $\theta_{uk} \sim \Gam(a_0, b_0)$.
  \end{itemize}
  \item For each item $i$ in $\{1,\cdots,I\}$:
  \begin{itemize}
    \item For each component $k$, draw $\beta_{ik} \sim \Gam(c_0, d_0)$.
  \end{itemize}
  \item For each user and item:
  \begin{itemize}
    \item Draw the observation $y_{ui} \sim
    \text{Poisson}(\mbtheta_u^\top\mbbeta_i)$.
  \end{itemize}
\end{enumerate}

\begin{figure}[htbp]
\includegraphics{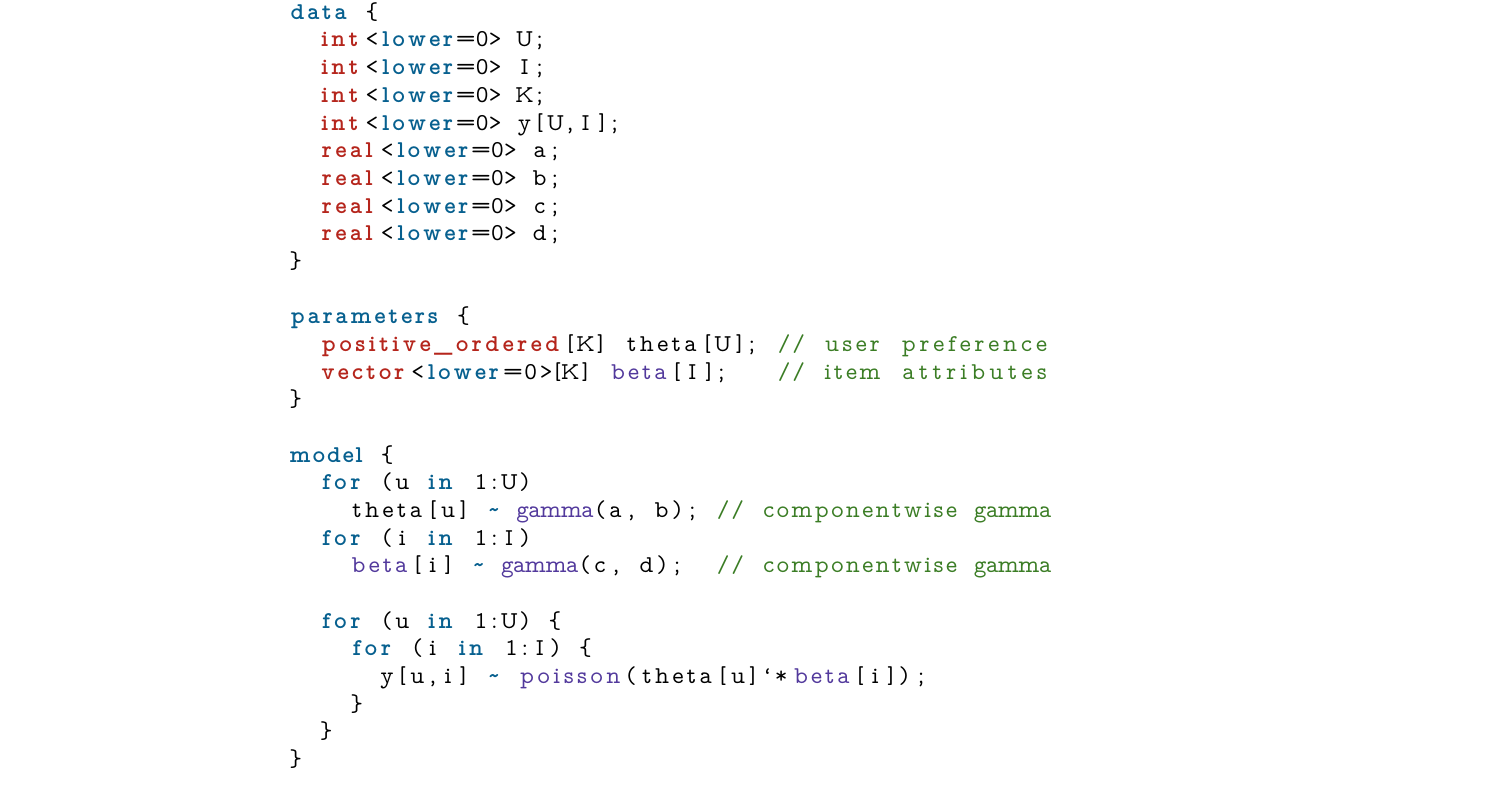}
  \caption{Stan code for the Gamma Poisson non-negative matrix factorization model.}
  \label{fig:code_nmf_pf}
\end{figure}

A potential downfall of this model is that it is not uniquely identifiable:
swapping rows and columns of $\mbtheta$ and $\mbbeta$ give the same inner
product. One way to contend with this is to constrain either vector to be an
 ordered vector during inference. We constrain each $\mbtheta_u$
vector in our model in this fashion. Stan code is in 
Figure\nobreakspace \ref {fig:code_nmf_pf}. We set $K=10$ and all the Gamma hyper-parameters to 1 in our
experiments.

\subsection{Non-negative Matrix Factorization: Dirichlet Exponential Poisson Model}
\label{app:dir_exp}

Another model for discrete data is a Dirichlet Exponential model. The
Dirichlet enforces uniqueness while the exponential promotes sparsity. This
is a non-conjugate model that does not appear to have been studied in the
literature.

The generative process for a Dirichlet Exponential model with $K$ factors is
\begin{enumerate}
  \item For each user $u$ in $\{1,\cdots,U\}$:
  \begin{itemize}
    \item Draw the $K$-vector $\mbtheta_{u} \sim \text{Dir}(\mb{\alpha}_0)$.
  \end{itemize}
  \item For each item $i$ in $\{1,\cdots,I\}$:
  \begin{itemize}
    \item For each component $k$, draw $\beta_{ik} \sim \text{Exponential}
    (\lambda_0)$. \end{itemize}
  \item For each user and item:
  \begin{itemize}
    \item Draw the observation $y_{ui} \sim
    \text{Poisson}(\mbtheta_u^\top\mbbeta_i)$.
  \end{itemize}
\end{enumerate}

\begin{figure}[htbp]
\includegraphics{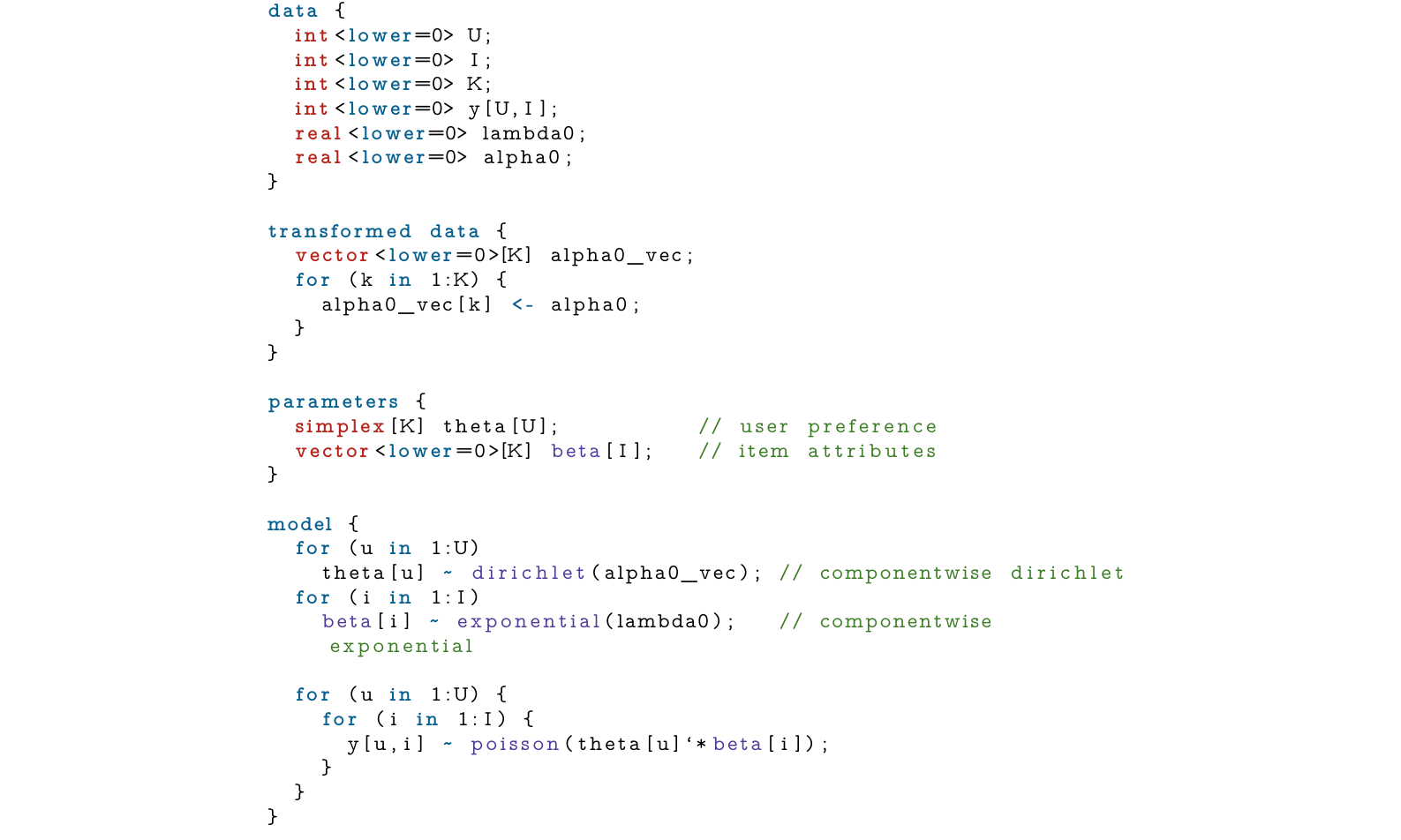}
  \caption{Stan code for the Dirichlet Exponential non-negative matrix factorization
  model.}
  \label{fig:code_nmf_dir_exp}
\end{figure}

Stan code is in Figure\nobreakspace \ref {fig:code_nmf_dir_exp}. We set $K=10$, $\alpha_0 = 1000$
for each component, and $\lambda_0 = 0.1$. With this configuration of
hyper-parameters, the
factors $\mbbeta_i$ appear sparse.

\subsection{Gaussian Mixture Model}
\label{app:gmm}
\glsreset{GMM}

The \gls{GMM} is a celebrated probability model \citep{bishop2006pattern}. 
We use it to group a dataset of
natural images based on their color histograms. We build a high-dimensional \gls
{GMM} with a Gaussian prior for the mixture means, a lognormal prior for the
mixture standard deviations, and a Dirichlet prior for the mixture components.

Represent the images as $\mby = y_{1:N}$ where each $y_n$ is $D$-dimensional and
there are $N$ observations. The likelihood for the images is
\begin{align*}
  p(\mby \mid \mbtheta, \mbmu, \mbsigma)
  &=
  \prod_{n=1}^N
  \sum_{k=1}^K
  \theta_k
  \prod_{d=1}^D
  \cN(y_{nd}\mid\mu_{kd},\sigma_{kd})
\end{align*}
with a Dirichlet prior for the mixture proportions
\begin{align*}
  p(\mbtheta)
  &=
  \text{Dir}(\mbtheta\,;\, \mb{\alpha}_0),
\end{align*}
a Gaussian prior for the mixture means
\begin{align*}
  p(\mbmu)
  &=
  \prod_{k=1}^D
  \prod_{d=1}^D
  \cN(\mu_{kd}\,;\, 0, 1)
\end{align*}
and a lognormal prior for the mixture standard deviations
\begin{align*}
  p(\mbsigma)
  &=
  \prod_{k=1}^D
  \prod_{d=1}^D
  \text{logNormal}(\sigma_{kd}\,;\, 0, 1)
\end{align*}

The dimension of the color histograms in the image\textsc{clef} dataset is $D =
576$. This is a concatenation of three $192$-length histograms, one for each
color channel (red, green, blue) of the images.

We scale the image histograms to have zero mean and unit variance. Setting
$\alpha_0$ to a small value encourages the model to use fewer components to
explain the data. Larger values of $\alpha_0$ encourage the model to use all $K$
components. We set $\alpha_0=1\,000$ in our experiments.

\begin{figure}[htbp]
\includegraphics{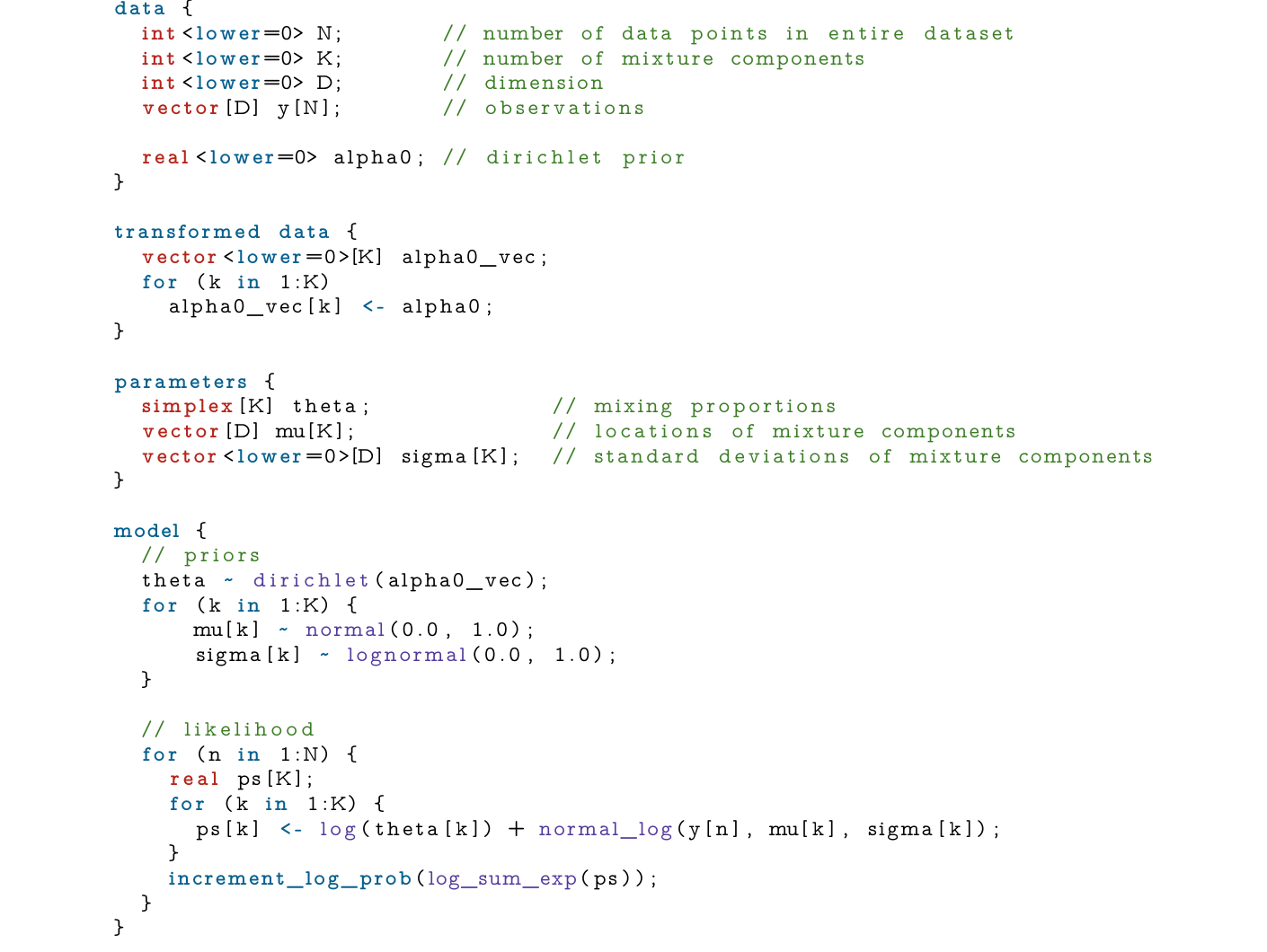}
  \caption{\gls{ADVI} Stan code for the \gls{GMM} example.}
  \label{fig:code_gmm_diag}
\end{figure}

\begin{figure}[htbp]
\includegraphics{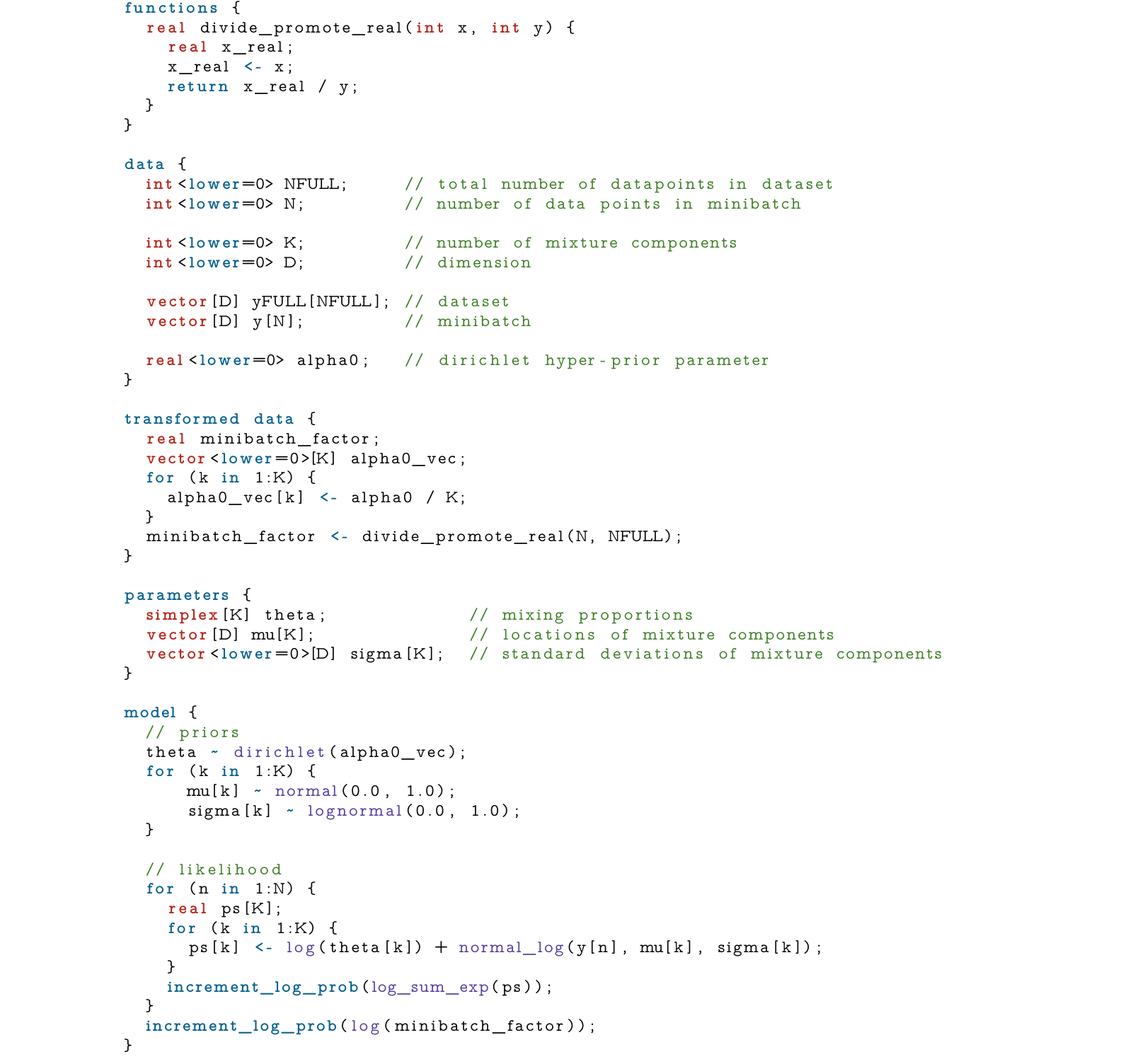}
  \caption{\gls{ADVI} Stan code for the \gls{GMM} example, with stochastic
  subsampling of the dataset.}
  \label{fig:code_gmm_diag_adsvi}
\end{figure}

\gls{ADVI} code is in Figure\nobreakspace \ref {fig:code_gmm_diag}. The stochastic data subsampling
version of the code is in Figure\nobreakspace \ref {fig:code_gmm_diag_adsvi}.

\glsreset{PPCA}
\subsection{Probabilistic Principal Component Analysis with Automatic Relevance
Determination}
\label{app:ppca_ard2}

\Gls{PPCA} is a Bayesian extension of classical principal component analysis
\citep{bishop2006pattern}. The generative process is straightforward.
Consider a dataset of $\mbx=x_{1:N}$ where each $x_n$ is $D$-dimensional.
Let $M$ be the dimension of the subspace we seek.

First define a set of latent
variables $\mbz=z_{1:N}$ where each $z_n$ is $M$-dimensional. Draw each
$z_n$ from a standard normal
\begin{align*}
  p(\mbz) &= \prod_{n=1}^N \cN(z_n\:;\:\mb0,\mbI).
\end{align*}
Then define a set of principal components $\mbw=w_{1:D}$ where each $w_d$
is $M$-dimensional. Similarly, draw the principal components from a standard
normal
\begin{align*}
  p(\mbw) &= \prod_{d=1}^D \cN(w_d\:;\:\mb0,\mbI).
\end{align*}
Finally define the likelihood through an inner product as
\begin{align*}
  p(\mbx \mid \mbw,\mbz,\sigma)
  &=
  \prod_{n=1}^N \cN(x_n\:;\:\mbw z_n,\sigma\mbI).
\end{align*}
The standard deviation $\sigma$ is also a latent variable. Place a lognormal
prior on it as
\begin{align*}
  p(\sigma) &= \text{logNormal}(\sigma\:;\:0,1).
\end{align*}

\begin{figure}[htbp]
\includegraphics{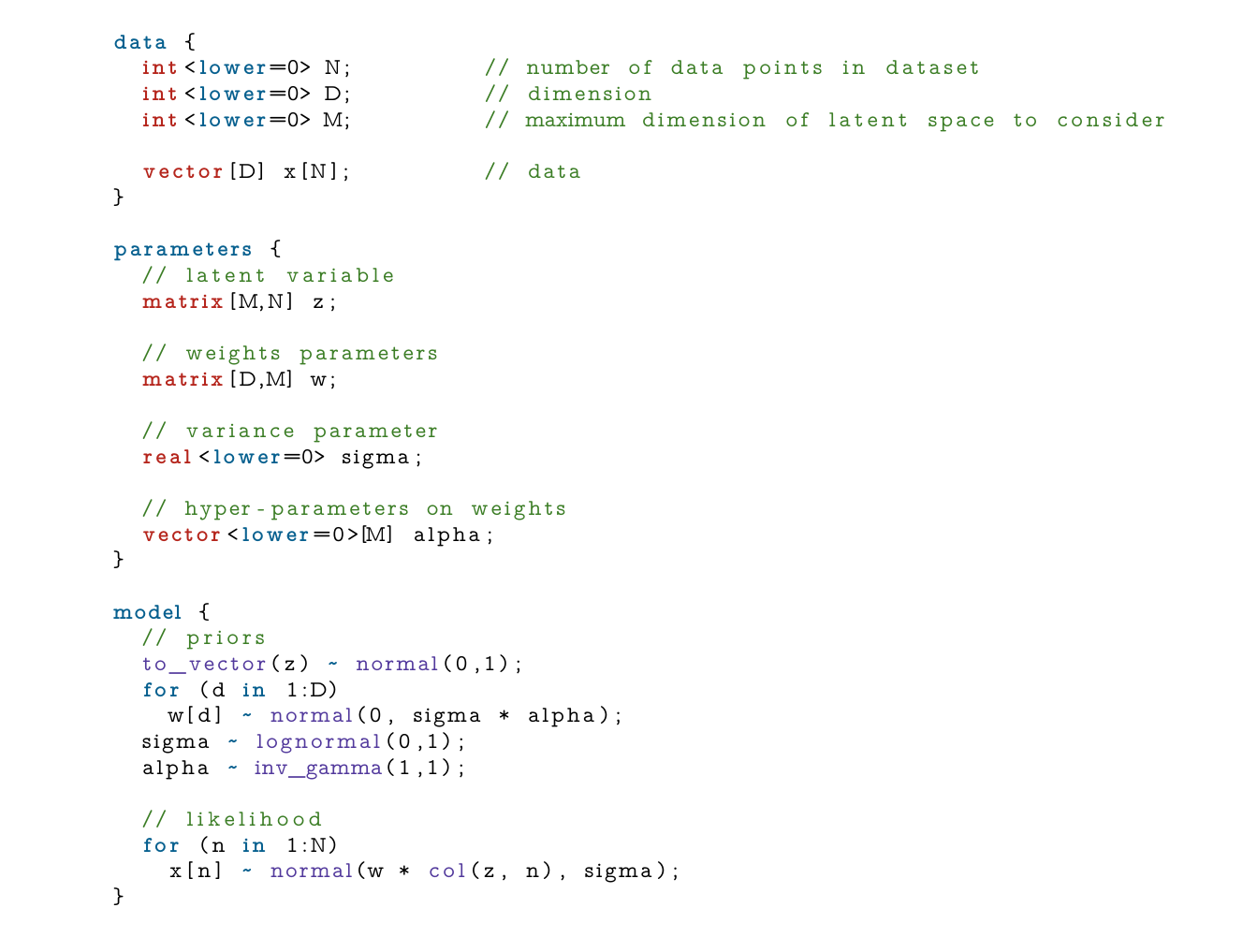}
  \caption{\gls{ADVI} Stan code for the \gls{PPCA} with \gls{ARD}.}
  \label{fig:code_ppca_ard2}
\end{figure}

We extend \gls{PPCA} by adding an \gls{ARD} hierarchical prior. The extended
model introduces a $M$-dimensional vector $\mbalpha$ which chooses which
principal components to retain. ($M < D$ now represents the maximum number of
principal components to consider.) The extended extends the above by
\begin{align*}
  p(\mbalpha) &= \prod_{m=1}^M \text{InvGamma}(\alpha_m\:;\:1,1)\\
  p(\mbw\mid\mbalpha) &= \prod_{d=1}^D \cN(w_d\:;\:\mb0,\sigma \diag
  (\mbalpha))\\
  p(\mbx \mid \mbw,\mbz,\sigma)
  &=
  \prod_{n=1}^N \cN(x_n\:;\:\mbw z_n,\sigma\mbI).
\end{align*}
\gls{ADVI} code is in Figure\nobreakspace \ref {fig:code_ppca_ard2}.

\glsreset{SUP-PPCA}
\subsection{Supervised Probabilistic Principal Component Analysis with Automatic Relevance
Determination}
\label{app:sup_ppca_ard2}

\begin{figure}[htbp]
\includegraphics{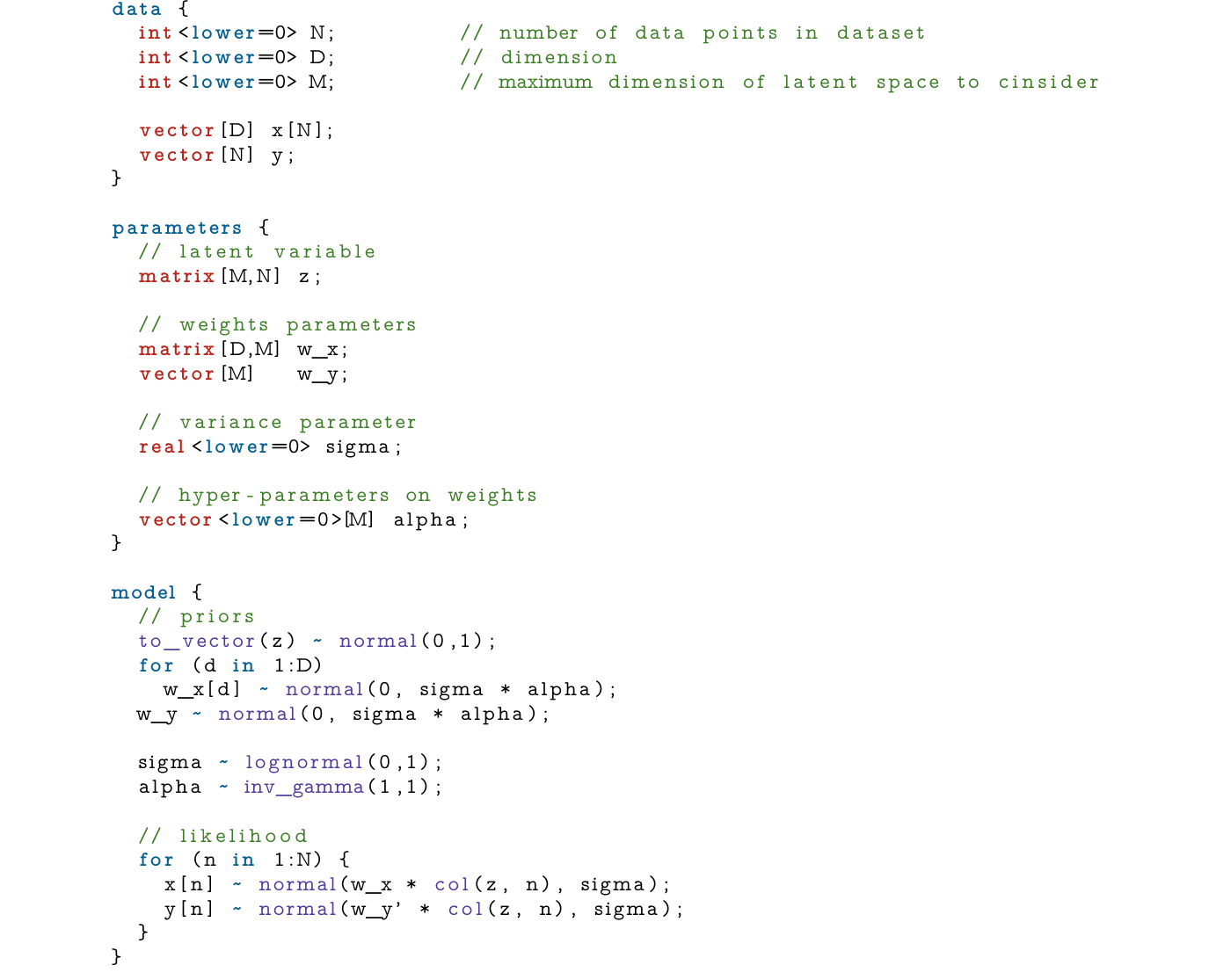}
  \caption{\gls{ADVI} Stan code for the \gls{SUP-PPCA} with \gls{ARD}.}
  \label{fig:code_sup_ppca_ard2}
\end{figure}

\Gls{SUP-PPCA} augments \gls{PPCA} by regressing a vector of observed random
variables $y$ onto the principal component subspace. The idea is to not only
find a set of principal components that describe variation in the dataset
$\mbx$, but to also predict $y$. The complete model is
\begin{align*}
  p(\mbz) &= \prod_{n=1}^N \cN(z_n\:;\:\mb0,\mbI)\\
  p(\sigma) &= \text{logNormal}(\sigma\:;\:0,1)\\
  p(\mbalpha) &= \prod_{m=1}^M \text{InvGamma}(\alpha_m\:;\:1,1)\\
  p(\mbw_\mbx\mid\mbalpha) &= \prod_{d=1}^D \cN(w_d\:;\:\mb0,\sigma \diag
  (\mbalpha))\\
  p(w_y\mid\mbalpha) &= \cN(w_y\:;\:\mb0,\sigma \diag
  (\mbalpha))\\
  p(\mbx \mid \mbw_\mbx,\mbz,\sigma)
  &=
  \prod_{n=1}^N \cN(x_n\:;\:\mbw_\mbx z_n,\sigma\mbI)\\
  p(y \mid w_y,\mbz,\sigma)
  &=
  \prod_{n=1}^N \cN(y_n\:;\:w_y z_n,\sigma).
\end{align*}
\gls{ADVI} code is in Figure\nobreakspace \ref {fig:code_sup_ppca_ard2}.

\clearpage

\clearpage
\bibliographystyle{apa}
\bibliography{bib}

\end{document}